\documentclass[runningheads]{llncs}
\usepackage{graphicx}
\usepackage{tikz}
\usepackage{comment}
\usepackage{amsmath,amssymb} % define this before the line numbering.
\usepackage{color}
\usepackage{graphicx}
\usepackage{amsmath}
\usepackage{amssymb}
\usepackage{booktabs}
\usepackage{bm}
\usepackage{mathtools}
\usepackage{multirow}
\usepackage{wrapfig}
\usepackage{enumitem}
\usepackage{url}
\usepackage{subcaption}
\usepackage{array}
\usepackage{tabularx}
\definecolor{newlightblue}{RGB}{0,75,255}
\usepackage[pagebackref,breaklinks,colorlinks, urlcolor={newlightblue}, citecolor={newlightblue}]{hyperref} 

\usepackage[accsupp]{axessibility}  % Improves PDF readability for those with disabilities.

\newcommand{\mypar}[1]{\vspace{-1.5mm}\paragraph{{\bf #1}}}
\newcommand{\mysubsec}[1]{\vspace{-0.5mm}\subsection{#1}}
\newcommand{\apppar}[1]{\vspace{-3mm}\paragraph{{\bf #1}}}    % for appendix only
\newcommand{\appsubsec}[1]{\vspace{-4mm}\subsection{#1}}    % for appendix only

\begin{document}
\pagestyle{headings}
\mainmatter

\title{Learning Visual Styles \\ from Audio-Visual Associations} % Replace with your title

%******************

% \begin{comment}
\titlerunning{Learning Visual Styles from Audio-Visual Associations}
\author{Tingle Li\inst{1,3} \and
Yichen Liu\inst{1} \and \\
Andrew Owens\inst{2} \and
Hang Zhao\inst{1,3}}
\authorrunning{T. Li et al.}
\institute{$^{1}$Tsinghua University
$^{2}$University of Michigan  
$^{3}$Shanghai Qi Zhi Institute\\
\url{https://tinglok.netlify.com/files/avstyle}}
%******************
\maketitle

%%%%%%%%% ABSTRACT
\begin{abstract}
From the patter of rain to the crunch of snow, the sounds we hear often convey the visual textures that appear within a scene. In this paper, we present a method for learning visual styles from unlabeled audio-visual data. Our model learns to manipulate the texture of a scene to match a sound, a problem we term {\em audio-driven image stylization}. Given a dataset of paired audio-visual data, we learn to modify input images such that, after manipulation, they are more likely to co-occur with a given input sound. In quantitative and qualitative evaluations, our sound-based model outperforms label-based approaches. We also show that audio can be an intuitive representation for manipulating images, as adjusting a sound's volume or mixing two sounds together results in predictable changes to visual style.
\end{abstract}

\vspace{-4mm}
\begin{figure}
	\centering
	\includegraphics[width=\textwidth]{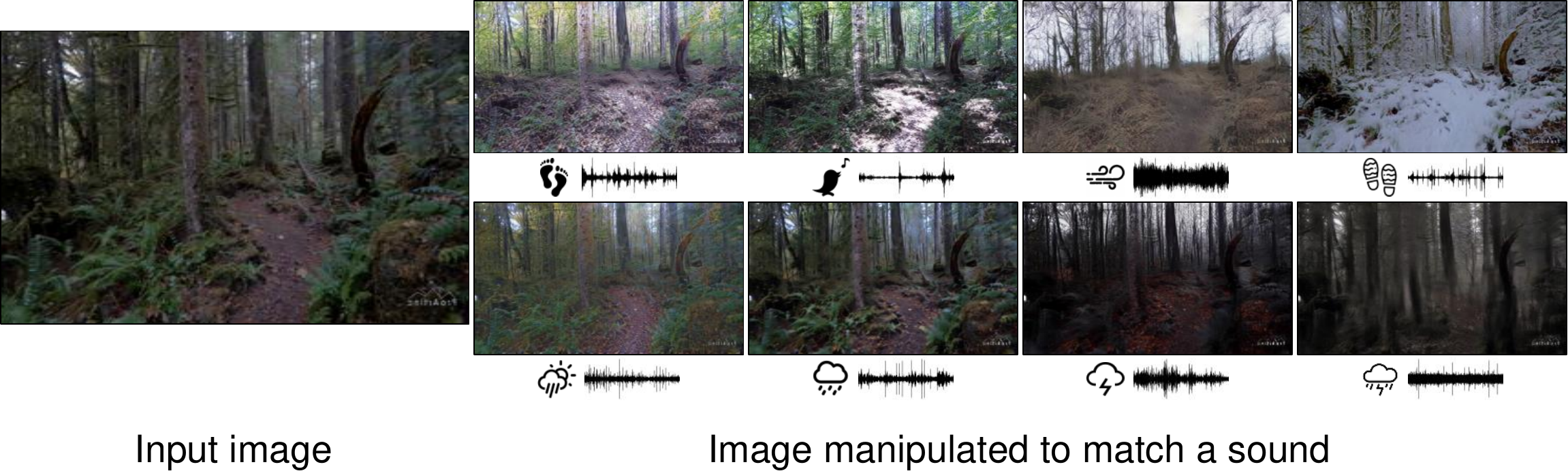}
	\caption{{\bf Audio-driven image stylization.} We manipulate the style of an image to match a sound. After training with an unlabeled dataset of egocentric hiking videos, our model learns visual styles for a variety of ambient sounds, such as light and heavy rain, as well as physical interactions, such as footsteps. } 
	\label{fig:teaser}
\end{figure}
\vspace{-5mm}
\section{Introduction}
\label{sec:intro}

Recent work has proposed a variety of methods for manipulating the style~\cite{tenenbaum2000separating,gatys2015neural} of an input image. In these methods, the desired style is specified using other example images~\cite{hertzmann2001image,gatys2015neural,johnson2016perceptual,huang2017arbitrary} and, more recently, through human language, such as through semantic labels, text, or scene graphs~\cite{reed2016generative,johnson2018image,bau2021paint,ramesh2021zero}. While this approach has been effective, it requires human-provided annotations and hence implicitly relies on a ``human in the loop." This supervision is often expensive to collect and may fail to capture important scene properties.

We propose to address these problems by learning stylization from {\em unlabeled audio-visual} data. Many scene properties, such as weather conditions, produce highly distinctive sights and sounds. Training a model to estimate visual information from audio requires it to identify these scene structures and, in the process, learn which visual textures are associated with a sound. 

Inspired by this idea, we introduce a model for performing {\em audio-driven} image stylization. Given an input image and a target sound, our model manipulates the textures within the image such that it better matches the sound, while preserving the image's structural content. Through this process, our model learns a variety of visual styles, each of which can be specified by a sound --- e.g., bird chirps and blue skies, crunching footsteps and snow,  rain and dark skies (Figure~\ref{fig:teaser}).

Audio naturally comes paired with visual data, and thus provides a free learning signal, complementing human-provided supervision like labels and text. It also conveys important distinctions between scenes that often may not be evident in pre-existing text or label sets. For example, asking a model to generate images depicting a ``rainy" scene can be ambiguous. Providing the sound of rain, on the other hand, specifies whether the rain is light or heavy, as well as whether the image is likely to contain dark, stormy skies. Finally, audio can be used as a natural representation for specifying image styles, as intuitive changes to the audio, such as adjusting the volume or mixing two sounds together, result in predictable visual changes.

Our model combines conditional generative adversarial networks \cite{adversarial_loss} and contrastive learning \cite{NCE}, following the recent approach of Park et al.~\cite{park2020contrastive}. We use an audio-visual discriminator to determine whether the generated image and target audio are likely to co-occur, with the goal of converting the source image's style. We also use an multi-scale patch-wise structure discriminator \cite{park2020contrastive} that maximizes the mutual information between the source and generated images in order to preserve the structural content of the scene. We train the model on a dataset of egocentric hiking videos collected from the internet.

After training, our model can manipulate images to match a variety of visual styles, each specified using sound. Through quantitative evaluations and human perceptual studies, we demonstrate the effectiveness of our model's ability to stylize images. We also provide qualitative results showing how that straightforwardly modifying the audio, by mixing it or changing its volume, leads to corresponding changes in image style. Through our evaluations, we show:

\def\labelitemi{\textbullet }
\begin{itemize}[topsep=0pt, noitemsep, leftmargin=26pt]
\item Unlabeled audio provides supervision for learning visual styles.
\item Our proposed model learns to perform audio-driven stylization from in-the-wild audio-visual data. 
\item Adjusting the volume of a sound or mixing it with other sounds lead to predictable changes in image style.
\end{itemize}
\section{Related Work} 
\label{sec: related_work}
\paragraph{\bf Image translation.}
Paired image translation~\cite{isola2017image} frames the image prediction problem as a straightforward supervised learning task, which corresponding input and target images. Unpaired image translation~\cite{adversarial_loss,zhu2017unpaired,yi2017dualgan,kim2017learning,park2020contrastive} learns to transform images between two different domains, without ground-truth correspondences. We take inspiration from work~\cite{laffont2014transient} that manipulates the global appearance of a scene, such as through labels indicating the desired weather. A variety of methods have been proposed generate or manipulate images based on text~\cite{reed2016generative,dong2017semantic,nam2018text,bau2021paint,ramesh2021zero}. In particular, our approach is closely related to Fu et al.~\cite{fu2021language}, which stylizes images based on text. However, text- and label-based methods either require ``weak" supervision~\cite{mahajan2018exploring,radford2021learning} from humans (\textit{e.g.}, paired text and images from webpages) or explicit image descriptions (\textit{e.g.}, text describing an image as a line drawing)~\cite{wu2020describing}. These descriptions may not capture the full range of image styles, and it requires significant effort from humans (including {\em implicit} effort through weak supervision). Our approach, by contrast, uses audio to learn styles, without any form of human labeling. It therefore provides a {\em complementary} learning signal to text and labels.

\mypar{Audio-visual correspondence.}
Audio and visual signals  naturally co-occur when they are recorded as video. In order to leverage this natural correspondence, researchers have introduced various tasks, such as representation learning \cite{de1994learning,ngiam2011multimodal,arandjelovic2017look,korbar2018cooperative,owens2018audio,morgado2021audio}, source separation \cite{zhao2018sound,zhao2019sound,ephrat2018looking,gao2018learning}, audio source grounding \cite{chen2021localizing,harwath2018jointly}, audio spatialization \cite{gao20192,morgado2018self,yang2020telling}, visual speech recognition \cite{afouras2018deep}, and scene classification \cite{chen2020vggsound,gemmeke2017audio}. Inspired by these works that use audio-visual correspondence, we propose a novel task termed audio-driven image stylization, aiming to conduct image translation using sounds like birds chirping, rain and footsteps.

\mypar{Audio-visual synthesis.}
A variety of methods have been proposed for synthesizing images from sound or vice versa. One line of work has generated sounds from video, such as impact sounds \cite{owens2016visually,zhang2017generative}, natural sounds~\cite{zhou2018visual,iashin2021taming}, or human speech \cite{hu2021neural,prajwal2020learning}. Another line of work has created models that synthesize images from sound, such as by generating talking heads~\cite{chung2017you,zhou2019talking,prajwal2020lip}, pose~\cite{shlizerman2018audio,gan2020music,levine2010gesture,ginosar2019learning}, synchronizing rigid body animations with contact sounds~\cite{langlois2014inverse}, estimating depth from ambient sound~\cite{chen2021structure}, predicting future video frames~\cite{chatterjee2020sound2sight}. Unlike these works, we concentrate on restyling plausible images using the source image and natural sounds. In concurrent work, Lee et al.~\cite{lee2021sound} used sound to guide a text-based image manipulation method based on CLIP~\cite{radford2021learning}. In contrast, our model learns image styles solely from unlabeled audio-visual data. 
\section{Audio-driven Image Stylization}
\label{sec:method}
\begin{wrapfigure}[12]{r}{0.7\textwidth}
  \begin{center}
  \vspace{-11.8mm}
    \includegraphics[width=0.7\textwidth]{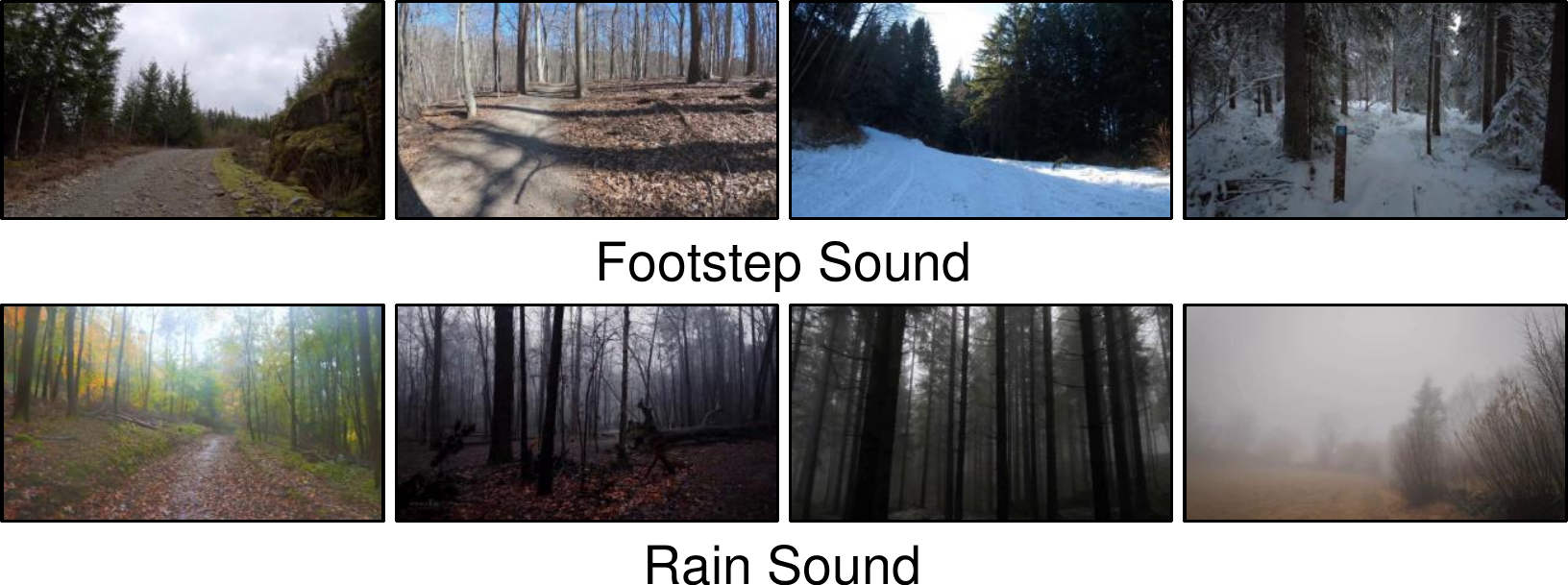}
  \end{center}
  \vspace{-5mm}
  \caption{Categories can fail to convey subtle distinctions between events. We show frames whose corresponding sounds were classified as {\em footstep} or {\em rain}~\cite{plakal2020yamnet,gemmeke2017audio}.}
  \label{fig:why_audio}
\end{wrapfigure}

We take inspiration from the fact that audio can convey distinctions that may not be obvious from semantic categories. For example, consider the  images shown in Figure~\ref{fig:why_audio}. While these videos have the same category ({\em e.g.}, rain), their visual style significantly varies ({\em e.g.}, heavy or light rain). This distinction, however, is easily captured by the corresponding sound. We propose {\em audio-driven} image stylization (ADIS) as a novel multi-modal generation task for learning these styles. 

We pose this problem as learning a mapping from a source image domain $\mathcal{X}$ to a target domain $\mathcal{Y}$ using an input sound from the audio domain $\mathcal{A}$. To achieve this goal, we propose a self-supervised learning approach that can be trained on unpaired videos. This can be accomplished through two distinct training objectives.

\mypar{Texture conversion via adversarial training.}
We introduce an audio-visual adversarial objective that discriminates whether an image is co-occurred with a given audio. Under this training scheme, the generated image is encouraged to match the target audio. Specifically, the generator $G$ consists of two components, an encoder $G_\text{enc}$ followed by a decoder $G_\text{dec}$. For a given dataset of unpaired image instances $X=\{\bm{x}\in\mathcal{X}\}$, $Y=\{\bm{y}\in\mathcal{Y}\}$, and the audios $A_Y=\{\bm{a}_Y\in\mathcal{A}\}$ corresponding to $Y$, $G_\text{enc}$ and $G_\text{dec}$ are applied sequentially to generate the output image $\hat{\bm{y}}=G_\text{dec}(\text{concat}(G_\text{enc}(\bm{x}), f(\bm{a}_Y)))$, where $f$ is a audio feature extractor. 

The audio-visual adversarial loss \cite{adversarial_loss} is then applied to increase the association between $\hat{\bm{y}}$ and $\bm{a}_Y$:
\begin{equation}\label{eq: adversarial_loss}
\begin{split}
    \mathcal{L}_{\text{GAN}}(G_{X\to Y},D_Y)= \text{ }&\mathbb{E}_{\bm{y}\sim Y}\log{D(\bm{y},\bm{a}_Y)} + \\
    &\mathbb{E}_{\bm{x}\sim X}\log{(1-D(G(\bm{x},f(\bm{a}_Y)),\bm{a}_Y))}
\end{split}
\end{equation}
where $D$ is the discriminator. In our model, $D$ performs early fusion, where the spectrogram of $\bm{a}_Y$ is directly concatenated to $\hat{\bm{y}}=G(\bm{x},\bm{a}_Y)$ before feeding into $D$. We empirically found that this fusion strategy yields better results in terms of visual quality.

\begin{figure}[t]
	\centering
	\includegraphics[width=\textwidth]{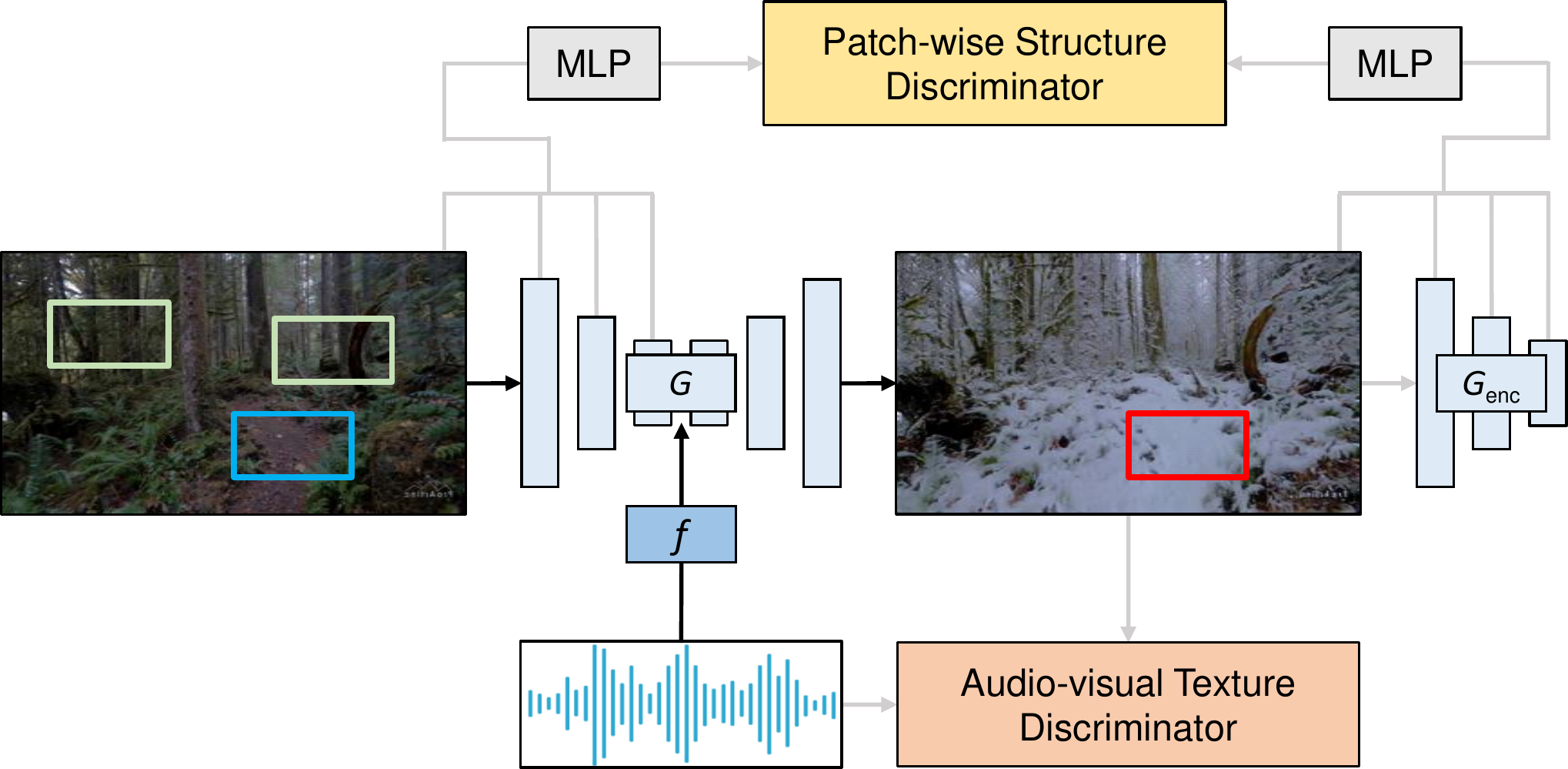}
	\caption{{\bf Model architecture}. The multi-scale patch-wise structure discriminator \cite{park2020contrastive} is used to preserve the scene structure, while the audio-visual texture discriminator is used to convert the scene texture. This is an example where sunny forest is converted to snowy counterpart. The \textcolor{red}{\textbf{generated snow patch}} should match its corresponding \textcolor[rgb]{0,0.69,0.94}{\textbf{input dirt patch}}, in comparison to \textcolor[rgb]{0.77,0.88,0.70}{\textbf{other random patches}}. Note that the MLP component is not used during inference.}
	\label{fig:model}
\end{figure}

\mypar{Structure preservation via contrastive learning.}
In this task, a successfully restyled image should be equipped with the texture that can be interpreted by the target audio, while fully preserving the structure of the source image. However, both information, \textit{i.e.}, texture and structure information, are inherently entangled within the learned feature, and adversarial training can only convert texture. One trivial solution could be that we get the same image for any inputs. Therefore, as shown in Figure \ref{fig:model}, we introduce the second training objective based on noise contrastive estimation (NCE) \cite{NCE}, which aims to preserve structure information by establishing mutual correspondence between the source and generated images, $\bm{x}$ and $\hat{\bm{y}}$ respectively. Note that this training objective is only employed to the encoder network $G_\text{enc}$, which is a multi-layer convolutional network that transforms the source image into feature stacks at each layer. In this way, we encourage $G_\text{enc}$ to abandon the texture of the source image while preserving the structure, and then the job of the decoder network $G_\text{dec}$ is to integrate the target texture to the source image. 

Given a ``query" vector $\bm{q}$, the objective in contrastive learning is to optimize the probability of selecting the corresponding ``positive" sample $\bm{v}^+$ among $N$ ``negative" samples $\bm{v}^-$. The query, positive and $N$ negatives are mapped to $M$-dimensional vectors by a MLP, $\textit{i.e.}$, $\bm{q},\bm{v}^+\in\mathbb{R}^M$ and $\bm{v}^-\in\mathbb{R}^{N\times M}$. This problem setting can be expressed as a multi-classification task with $N+1$ classes:
\begin{equation}\label{eq: NCE_expression}
    \ell(\bm{q},\bm{v}^+,\bm{v}^-)=-\log \left(\frac{\exp(\bm{q}\cdot\bm{v}^{+}/{\tau})}{\exp(\bm{q}\cdot\bm{v}^{+}/{\tau})+\Sigma_{n=1}^N\exp(\bm{q}\cdot\bm{v}^{-}_{n}/{\tau})} \right) 
\end{equation}
where $\bm{v}^-_n$ denotes the n-th negative sample and $\tau$ is a temperature parameter, as suggested in SimCLR \cite{SimCLR}, that scales the similarity distance between $\bm{q}$ and other samples. The cross-entropy term in Eq.\eqref{eq: NCE_expression} represents the probability of matching $\bm{q}$ with the corresponding positive sample $\bm{v}^+$. Thus, iteratively minimizing the negative log-cross-entropy is equivalent to establishing mutual correspondence between the query and sample spaces.

In our task, we draw the $N+1$ positive/negative samples from the source image $\bm{x}\in X$, and the query $\bm{q}$ is selected from the generated image $\hat{\bm{y}}$. From Figure \ref{fig:model}, it can be seen that the selected samples are ``patches" that capture local information among the image features. This setup is motivated by the logical assumption that the global correspondence between $\bm{x}$ and $\hat{\bm{y}}$ is determined by the local, \textit{i.e.}, patch-wise, correspondences. 

Since the encoder $G_\text{enc}$ is a multi-layer convolutional network that maps $\bm{x}$ into feature stacks after each layer, we choose $L$ layers and pass their feature stacks through a small MLP network $P$. The output of $P$ is $P(G^l_\text{enc}(\bm{x}))=\{\bm{v}_l^1,...,\bm{v}_l^N,\bm{v}_l^{N+1}\}$, where $l\in\{1,2...,L\}$ denotes the index of the chosen encoder layers and $G_\text{enc}^l(\bm{x})$ is the output feature stack of the $l$-th layer. Similarly, we can obtain the query set by encoding the generated spectrogram $\hat{\bm{y}}$ into $\{\bm{q}_l^1,...,\bm{q}_l^N,\bm{q}_l^{N+1}\}=P(G_\text{enc}^l(\bm{\hat{y}}))$. Now we let $\bm{v}_l^n\in\mathbb{R}^M$ and $\bm{v}_l^{(N+1)\text{\textbackslash}n}\in \mathbb{R}^{N\times M}$ denote the corresponding positive sample and the $N$ negative samples, respectively, where $n$ is the sample index and $M$ is the channel size of $P$. By referring to Eq.\eqref{eq: NCE_expression}, our second training objective can be expressed as:
\begin{equation}\label{eq: NCE_objective}
    \mathcal{L}_\text{NCE}(G_\text{enc},P,X)=\mathbb{E}_{\bm{x}\sim X} \sum_{l=1}^{L} \sum_{n=1}^{N+1} \ell(\bm{q}_l^n,\bm{v}_l^n,\bm{v}_l^{(N+1)\text{\textbackslash}n})
\end{equation}
which is the average NCE loss from all $L$ encoder layers.

\mypar{Overall objective.}
In addition to the two objectives discussed above, we have also employed an identity loss $\mathcal{L}_\text{identity}=\mathcal{L}_\text{NCE}(G_\text{enc},P,Y)$ which also leverages the NCE expression in Eq.\eqref{eq: NCE_objective}. By taking the NCE loss on the identity generation process, $\textit{i.e.}$, generating $\hat{\bm{y}}$ from $\bm{y}$, we are likely to prevent the generator from making unexpected changes. Now we can define our final training objective as:
\begin{equation}\label{eq: final_objective_loss}
\begin{split}
    \mathcal{L}_\text{final}= \text{ }&\mathcal{L}_{\text{GAN}}(G_{X\to Y},D_Y)+\lambda\mathcal{L}_\text{NCE}(G_\text{enc},P,X)+ \\ &\mu\mathcal{L}_\text{NCE}(G_\text{enc},P,Y)
\end{split}
\end{equation}
where $\lambda$ and $\mu$ are two parameters for adjusting the strengths of the NCE and identity loss.
\section{Experiments}
\label{sec: exp}

\begin{wrapfigure}[12]{r}{0.7\textwidth}
  \begin{center}
  \vspace{-11.8mm}
    \includegraphics[width=0.7\textwidth]{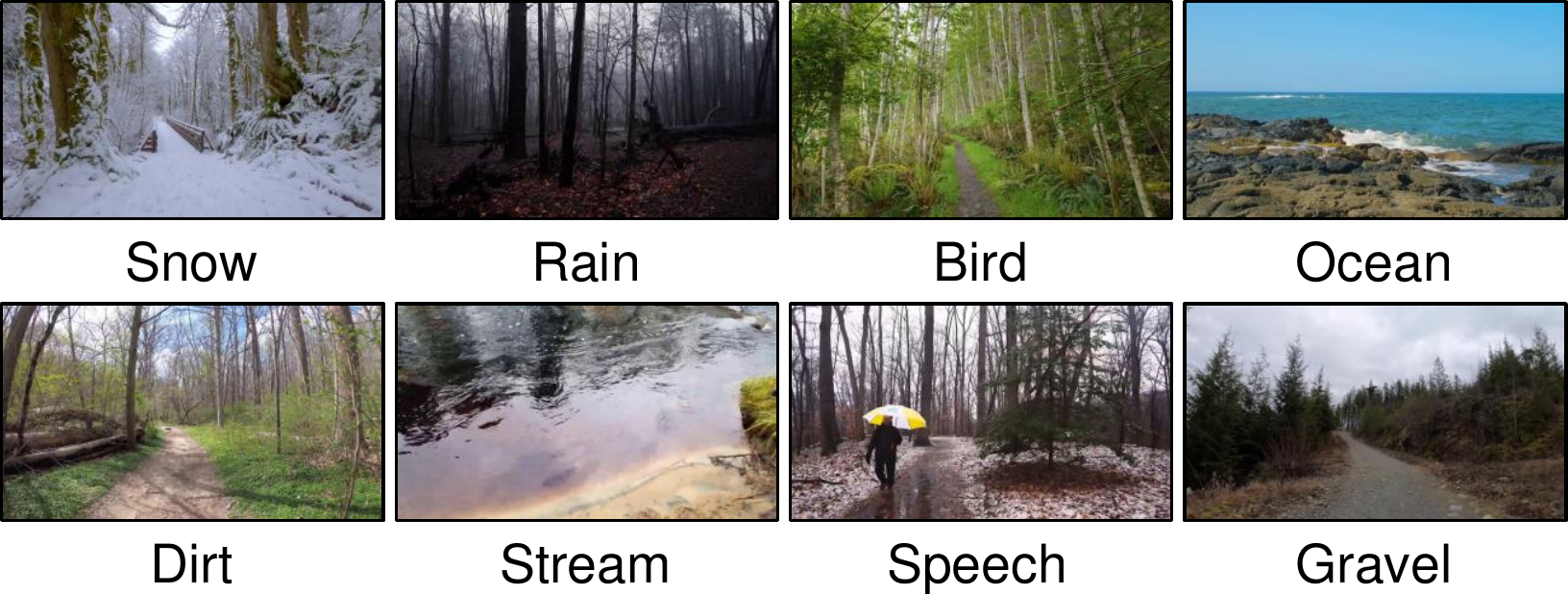}
  \end{center}
  \vspace{-5mm}
  \caption{{\bf Selected frames from the {\em Into the Wild} dataset.} We show example images corresponding to the top-1 categorical sounds deduced by a classifier \cite{plakal2020yamnet,gemmeke2017audio}.}
  \label{fig:example_frame}
\end{wrapfigure}

\mysubsec{Experimental Setup}
\paragraph{\bf Dataset.} We perform ADIS with two different datasets: \textit{Greatest Hits} and \textit{Into the Wild}. The former provides impact sounds from different materials, while the latter is a new dataset of egocentric hiking videos.
\begin{itemize}[topsep=0pt, noitemsep, leftmargin=*]
\item \textbf{\textit{Into the Wild} dataset}: We collect a new dataset to study the audio-visual associations that one would encounter on a hike (Figure~\ref{fig:example_frame}). These include sounds that are related to seasonal variations, rainfall,  animal vocalizations, and footsteps. We collect 94 untrimmed egocentric videos from YouTube, ranging from 1.5 to 130 minutes long (50 hours in total). 
We chose videos that only contain sounds naturally present in the scene ({\em e.g.,} no background music). See Appendix~\ref{sec: appendix_dataset} for more dataset details.

\item \textbf{The \textit{Greatest Hits} dataset} \cite{owens2016visually}: The \textit{Greatest Hits} dataset contains a drumstick hitting, scratching, and poking different objects in both indoor and outdoor scenes. There are 977 videos in total, including both indoor (64\%) and outdoor scenes (36\%). However, since this dataset was originally gathered for sound generation, each video more or less contains visual noise, making it challenging to perform ADIS. For example, ceramic bowls have different colors but the hitting sounds are similar across all bowls. It can be sometimes difficult for the model to determine the texture of a material with different colors. To alleviate this issue, we manually select some outdoor scene videos with less diverse backgrounds, such as dirt, water, gravel and grass.
\end{itemize}

\mypar{Network architecture.} The encoder and decoder of the GAN generator are 2D fully convolutional networks, with 9 layers of ResNet-based CNN bottlenecks \cite{johnson2016perceptual} in between. Except for the first CNN layer with a kernel size of 7 $\times$ 7, the others are 3 $\times$ 3, and the stride size is determined by whether downsampling is required. We used the PatchGAN architecture \cite{isola2017image} for the discriminator. A ResNet18 backbone \cite{he2016deep} is also used for extracting audio features before feeding them into the decoder of the GAN generator. Furthermore, before computing the NCE loss, we extract intermediate features from the encoder of the generator with five different scales, and then apply a 2-layer MLP with 256 units to map each feature.

\mypar{Training details.} For training efficiency, we devise the following pre-processing paradigm: \romannumeral1) before saving as images, each video is interpolated to 512$\times$512 scale and uniformly sampled 8 frames from it; \romannumeral2) each audio is randomly truncated or tiled to a fixed duration of 3 seconds, then converted to 16 kHz and 32-bit precision in floating-point PCM format; \romannumeral3) nnAudio \cite{cheuk2020nnaudio} is used for conducting a 512-point discrete Fourier transform with a frame length of 25 ms and a frame-shift of 10 ms. For the hyperparameters, both $\lambda$ and $\mu$ in Eq.\eqref{eq: final_objective_loss} are set to 0.5. We also employ random crop and horizontal flip as data augmentation. Our model is trained using the Adam optimizer \cite{kingma2014adam} with a batch size of 16 and an initial learning rate of $2 \times 10^{-4}$ over 50 epochs. Other training strategies are described in Appendix~\ref{sec: appendix_details}.

\mypar{Evaluation metrics.}
To get a better understanding of why audio is important, we quantitatively compare our model to several label-based baselines, using both objective and subjective metrics (see Appendix~\ref{sec: appendix_evaluation} for more evaluation details): 
\begin{itemize}[topsep=0pt, noitemsep, leftmargin=*]
    \item \textbf{Audio-visual Correspondence (AVC)}~\cite{arandjelovic2017look}: AVC measures the correlation between audio and image. In our case, we extract audio and visual features using OpenL3 \cite{cramer2019look}, a variant of L3-Net \cite{arandjelovic2017look} pre-trained on AudioSet \cite{gemmeke2017audio}, and then use those features to compute the average cosine similarity. A higher correlation is associated with a higher AVC score.   
    \item \textbf{Fréchet Inception Distance (FID)} \cite{heusel2017gans}: FID estimates the distribution of real and generated image activations using trained network and measures the divergence between them. A lower FID score indicates that real and generated images are more relevant.
    \item \textbf{Amazon Mechanical Turk (AMT)}: We use human participants to evaluate the audio-visual correlations (\textit{i.e.}, via a subjective evaluation). Each participant is asked to rank the quality of the correlation between a sound and the images generated by various methods. The scores range from 1 (indicating low correlation) to 4 (high correlation). 
    \item \textbf{Contrastive Language-Image Pretraining (CLIP)} \cite{radford2021learning}: CLIP is a network trained using contrastive learning to associate corresponding image and text pairs. In order to provide an additional evaluation metric that captures semantics, we use the keywords from the title of each video as text inputs to CLIP, then measure the text-image similarity. A higher CLIP score indicates a better correlation between a given text and image.
\end{itemize}

\mypar{Baselines.}
We adopt two label-based methods for comparison. For both of them, Word2Vec \cite{mikolov2013efficient} is used for generating the class embeddings, which is incorporated with the input image and serves as a textual condition. In addition, we create an image-conditioned baseline.
\begin{itemize}[topsep=0pt, noitemsep, leftmargin=*]
    \item \textbf{Class Pred.} \cite{plakal2020yamnet}: we use YAMNet, a state-of-the-art audio classification network \cite{hershey2017cnn} trained on AudioSet \cite{gemmeke2017audio}, to calculate the class logits. It is employed as an auto-labeling method to yield the semantic labels for all the audio clips.
    \item \textbf{Keyword}: Keyword is a human-labeling method in which each audio class is manually labeled with keywords from the video title, thereby conveying the information provided in the video metadata.
    \item \textbf{AdaIN}~\cite{huang2017arbitrary}: AdaIN is an image-conditioned arbitrary stylization method that incorporates the adaptive instance normalization to fuse the content image and the style one. It takes two images as input and restyles one to match the other. Note that the style image is picked at random from the video frames corresponding to the selected audio.
\end{itemize}

\begin{table}[t]
	\centering
	\caption{Evaluation results on the \textit{Into the Wild} dataset. The subjective AMT metric is presented with 95\% confidence intervals.}
	\setlength{\tabcolsep}{1.5mm}{
	\begin{tabular}{l c c c | c}
		\toprule
    		\multirow{2}*{\textbf{Method}} & \multicolumn{4}{c}{\textbf{Evaluation Metrics}} \\ \cmidrule{2-5} & AVC ($\uparrow$) & FID ($\downarrow$) & AMT ($\uparrow$) & CLIP ($\uparrow$)\\
		\midrule
		Target & 0.842 & / & / & 0.247\\
		\midrule
		Class Pred.~\cite{plakal2020yamnet} & 0.801 & 91.417 & 1.833 $\pm$ 0.042 & 0.228\\
		AdaIN~\cite{huang2017arbitrary} & 0.812 & 62.851 & 2.269 $\pm$ 0.044 & 0.232\\
		Keyword & 0.809 & 38.066 & 2.626 $\pm$ 0.045 & 0.236\\
		\midrule
		Ours & \textbf{0.820} & \textbf{34.139} & \textbf{3.273 \boldsymbol{$\pm$} 0.046} & \textbf{0.238}\\
		\bottomrule
	\end{tabular}}
	\label{tb:all-result}
\end{table}

\begin{table}[t]
% \scriptsize
	\centering
	\caption{AVC metric of specific scenes under our model and label-based baselines on the \textit{Into the Wild} dataset.}
	 \setlength{\tabcolsep}{1.35mm}{
	\begin{tabular}{l c c c}
		\toprule
		\multirow{2}*{\textbf{Method}} & \multicolumn{3}{c}{\textbf{Audio-visual Correspondence (\boldsymbol{$\uparrow$})}} \\ 
		\cmidrule{2-4} & Sunny-to-Rainy & Snowy-to-Sunny & Sunny-to-Snowy \\
		\midrule
		Class Pred.~\cite{plakal2020yamnet} & 0.819 & 0.796 & 0.793 \\
		Keyword & 0.827 & 0.802 & 0.808 \\
		\midrule
		Ours & \textbf{0.831} & \textbf{0.820} & \textbf{0.816} \\
		\bottomrule
	\end{tabular}}
	\label{tb:specific-result}
\end{table}

\mysubsec{Comparison to Baselines}
\paragraph{\bf Quantitative results.}
Since the diverse hitting and scratching sounds are not well-modeled by AudioSet~\cite{gemmeke2017audio}, which L3-Net~\cite{arandjelovic2017look} is trained on, we cannot meaningfully evaluate the \textit{Greatest Hits} with the AVC metric. As a result, we only provide quantitative results yielded from the \textit{Into the Wild} dataset. Table~\ref{tb:all-result} shows the quantitative comparisons between our model and label/image-conditioned baselines. For objective evaluation, our model outperforms three baselines across the AVC, FID, and CLIP metrics, suggesting that our model can generate more realistic images. In particular, our method outperforms AdaIN \cite{huang2017arbitrary}, despite the fact that AdaIN has already been pre-trained using ImageNet while ours is trained from scratch. We find that Keyword outperforms Class Pred., perhaps due to errors introduced by automatic labeling. Notably, Class Pred. contains 132 label classes from AudioSet, whereas Keyword only has 3 classes (sunny, snowy and rainy), which are all closely related to the scenes in \textit{Into the Wild}. We also observe that the CLIP metric for our model is on par with Keyword, which also indicates the benefit of using audio over labels. For human evaluation, we randomly select 1000 images from the test set, and ask participants to assess the level of the audio-visual correlation. It turns out that they consistently preferred our model's results, as shown in the penultimate column of Table~\ref{tb:all-result}, which is consistent with the objective evaluation results.

To gain a better understanding of our model's performance, we divide the entire test set into three categories: sunny, rainy, and snowy and report results on each subset. In this experiment, as shown in Table~\ref{tb:specific-result}, our model still holds the best performance compared to label-based baselines. Furthermore, we observe that when the target scene is sunny, the disparity between our model and Keyword (0.018) is larger than that of other scenes (0.004 \& 0.008). This may be because the ambient sounds in sunny forests are highly varied (\textit{e.g.}, crunching gravel/leave, birds chirping, \textit{etc}.).

\mypar{Qualitative results.}
We show qualitative results in Figure~\ref{fig:results} and provide additional results in the Appendix~\ref{sec: appendix_additional_results}. We note that all of the results are produced by a single model, \textit{i.e.}, through ``one-to-many" conversion. We observe that the AdaIN model sometimes cannot reliably preserve the input image's content (the first row of first input image). The Keyword model can generate plausible images that match the class of the target audio, but with apparent flaws when converting between the same scene categories (the second row of the second input image). For the YAMNet model, the generated images occasionally match the target images, but this does not happen in all cases. This may be because the success of a stylization is strongly dependent on whether the labels inferred by YAMNet are correct. Our model, by comparison, can capture the subtle distinctions within the same scene class. For example, our model can adjust the hue of the snow, when given a wind-and-footstep sound (which is not successfully captured by other models).

\begin{figure}[t]
	\centering
	\includegraphics[width=\linewidth]{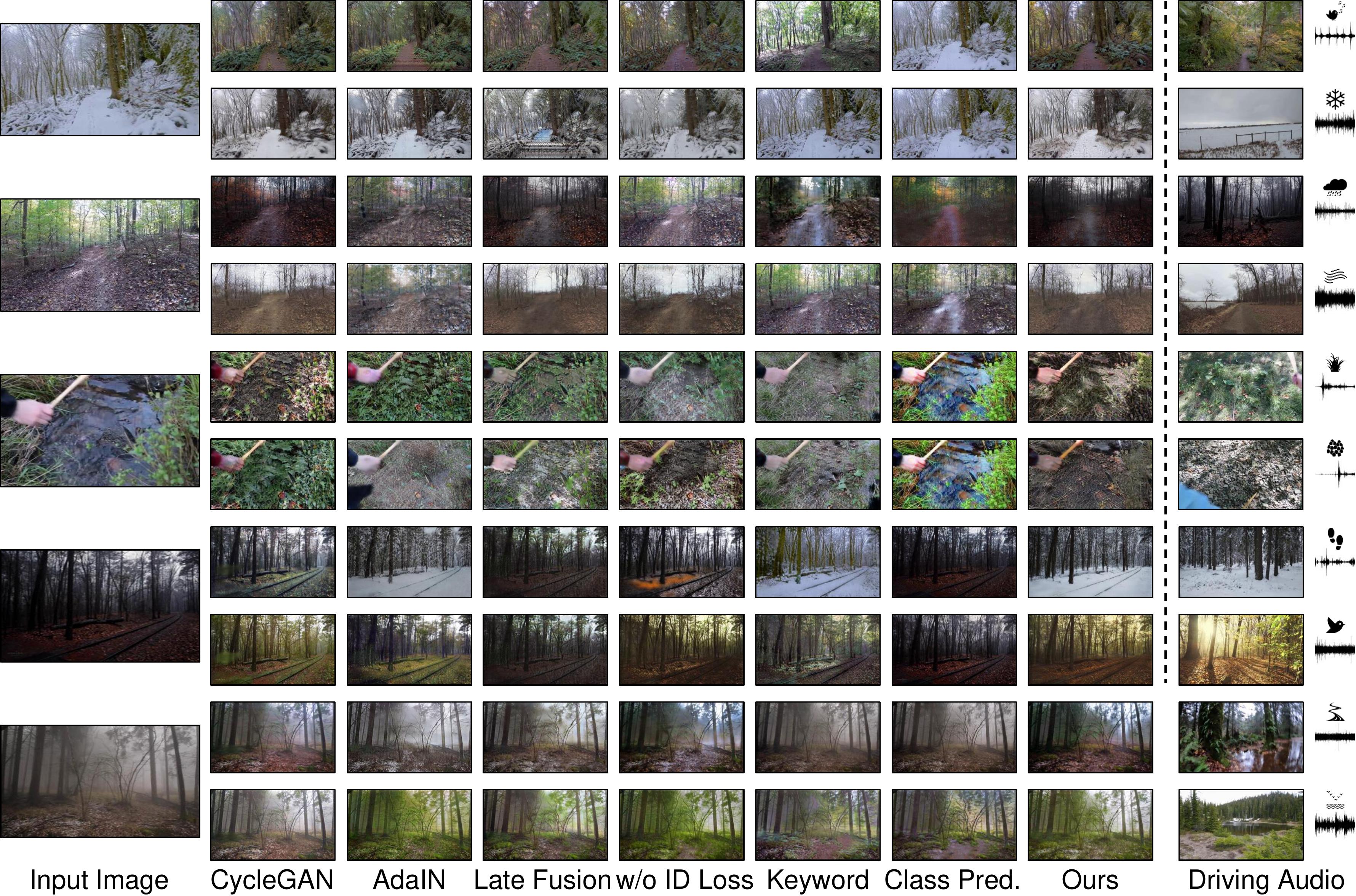}
    	\caption{Qualitative comparison of baselines, ablations, and our model on audio-visual texture conversion. For reference, we also show driving audios as well as their corresponding images in the last column.}
	\label{fig:results}
\end{figure}

\mysubsec{Ablation Study and Analysis}
We conduct an ablation study to test various settings and ablations of our model, summarized in Table~\ref{tb:ablations}. By default, we use the architecture and loss function above. We also try to use: \romannumeral1) the forward cycle-consistency loss \cite{zhu2017unpaired} instead of NCE loss, termed as CycleGAN; \romannumeral2) late fusion discriminator \cite{wang2020makes} to incorporate audio and visual features rather than early fusion one; \romannumeral3) without the identity loss; \romannumeral4) a pre-trained audio-visual self-supervised method, \textit{i.e.}, SeLaVi \cite{asano2020labelling}, as the initial weight for the audio network in addition to training from scratch. Besides, we show qualitative examples and additional pre-training comparisons in Figure~\ref{fig:results} and Appendix~\ref{sec: appendix_additional_results} respectively.

\mypar{NCE loss is a strong substitute for cycle-consistency loss.}
Our model employs NCE loss following CUT \cite{park2020contrastive}. As a baseline, cycle-consistency loss \cite{zhu2017unpaired} can also preserve the image structure. As shown in Table~\ref{tb:ablations}, our model achieves comparable results to its counterpart, CycleGAN, implying that it can generate realistic images like CycleGAN. Figure~\ref{fig:results} also shows some qualitative results that support this. Besides, CycleGAN involves the joint learning of two generators, while our model only requires one, which can reduce training time \cite{park2020contrastive}.

\begin{table}[t]
	\centering
	\caption{Quantitative results for ablations on \textit{Into the Wild} dataset.}
     \setlength{\tabcolsep}{3.6mm}{
	\begin{tabular}{l c c | c}
		\toprule
    	\multirow{2}*{\textbf{Ablation}} & \multicolumn{3}{c}{\textbf{Objective Evaluation}} \\ 
		\cmidrule{2-4} & AVC ($\uparrow$) & FID ($\downarrow$) & CLIP ($\uparrow$) \\
		\midrule
		CycleGAN~\cite{zhu2017unpaired} & 0.812 & 35.244 & 0.232 \\
		Late Fusion~\cite{wang2020makes} & 0.811 & 54.025 & 0.230 \\
		w/o ID Loss & 0.810 & 41.019 & 0.236 \\
		\midrule
		Ours & 0.820 & 34.139 & 0.238 \\
		+ Pre-training~\cite{asano2020labelling} & \textbf{0.822} & \textbf{32.882} & \textbf{0.242} \\
		\bottomrule
	\end{tabular}}
	\vspace{-3.2mm}
	\label{tb:ablations}
\end{table}

\mypar{Late fusion discriminators are more likely to collapse.}
In audio-visual learning, the late fusion architecture \cite{wang2020makes} is commonly used, in which two uni-modal encoders are employed to extract features, followed by a classifier (discriminator). We also take into account this architecture in ablations, with the results shown in Table~\ref{tb:ablations} and Figure~\ref{fig:results}. We find that leveraging this type of discriminator induces the model to collapse, which means the generator would eventually become too weak to sustain the image structure, resulting in unsatisfactory results.

\mypar{Identity loss helps to capture nuances.}
Given an image from the output domain, the identity loss \cite{zhu2017unpaired} pushes the generator to leave the image unchanged with our patch-based contrastive loss. We also test a variant without this loss, as depicted in Table~\ref{tb:ablations}. We find that the variation of the model without identity loss tends to has worse performance. We further investigate by presenting qualitative results in Figure~\ref{fig:results}. In the first row of the second example, in particular, when the conversion is from sunny to rainy forest, it is unsuccessful for the one without identity loss, whilst the one with succeeds. As a result, we propose that employing such a loss as a regularizer might be beneficial in capturing nuances, particularly when converting between similar landscapes, such as forest-to-forest and snow-to-snow conversions.

\mypar{Self-supervised pre-training improves stylization.}
We ask whether models pre-trained to solve audio-visual self-supervised learning tasks will result in performance gains. Table~\ref{tb:ablations} shows that fine-tuning our task using a pre-trained SeLaVi model~\cite{asano2020labelling} yields a small improvement.

\mysubsec{Audio Manipulation for Image Manipulation}
Sound provides a natural ``embedding space" for image manipulation, since intuitively manipulating the audio leads to corresponding changes in the images. We ask whether changing the volume of the sound or mixing two sounds together will result in corresponding visual changes. We also evaluate out-of-distribution images and audio.

\begin{figure}[t]
	\centering
	\includegraphics[width=\linewidth]{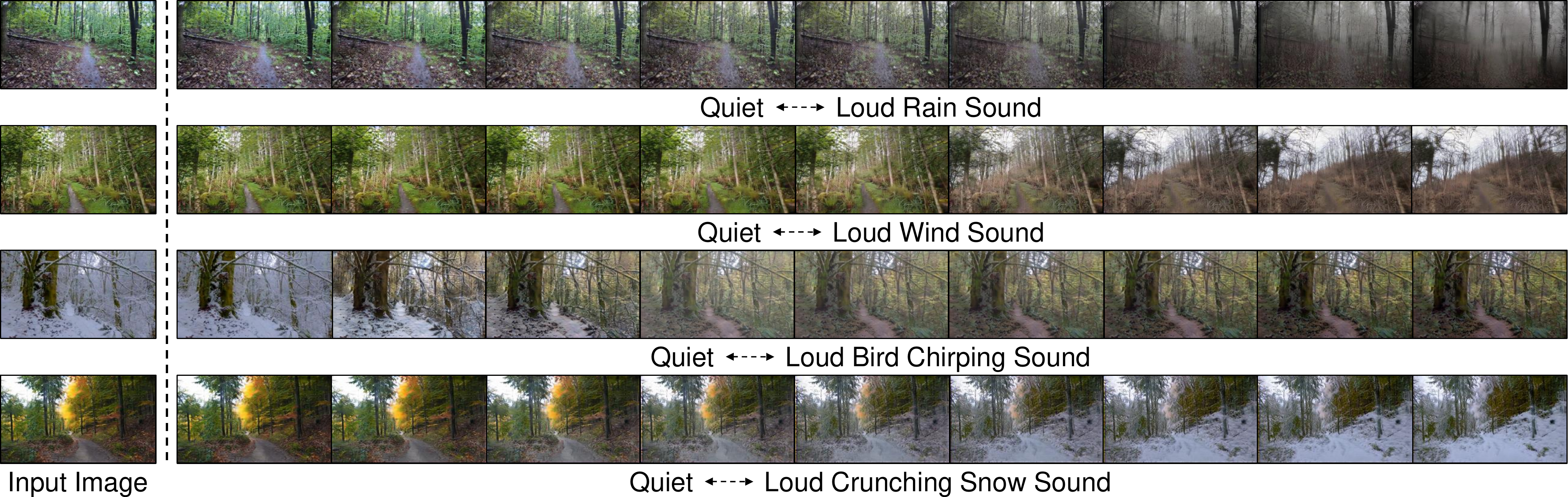}
	\caption{Qualitative results on image manipulation with increasing sound volumes.}
	\label{fig:audio_volume}
\end{figure}

\mypar{Changing sound volumes.}
A qualitative comparison using a sound at various volumes is shown in Figure~\ref{fig:audio_volume}. This is accomplished by simply rescaling the input waveform. Regardless of whether the input image is snowy or sunny forest, we observe that the texture in the image becomes more prominent as the sound gradually increases, indicating that our model implicitly learns to predict the prominence of the texture according to the volume.

\begin{figure}[t]
	\centering
	\includegraphics[width=\linewidth]{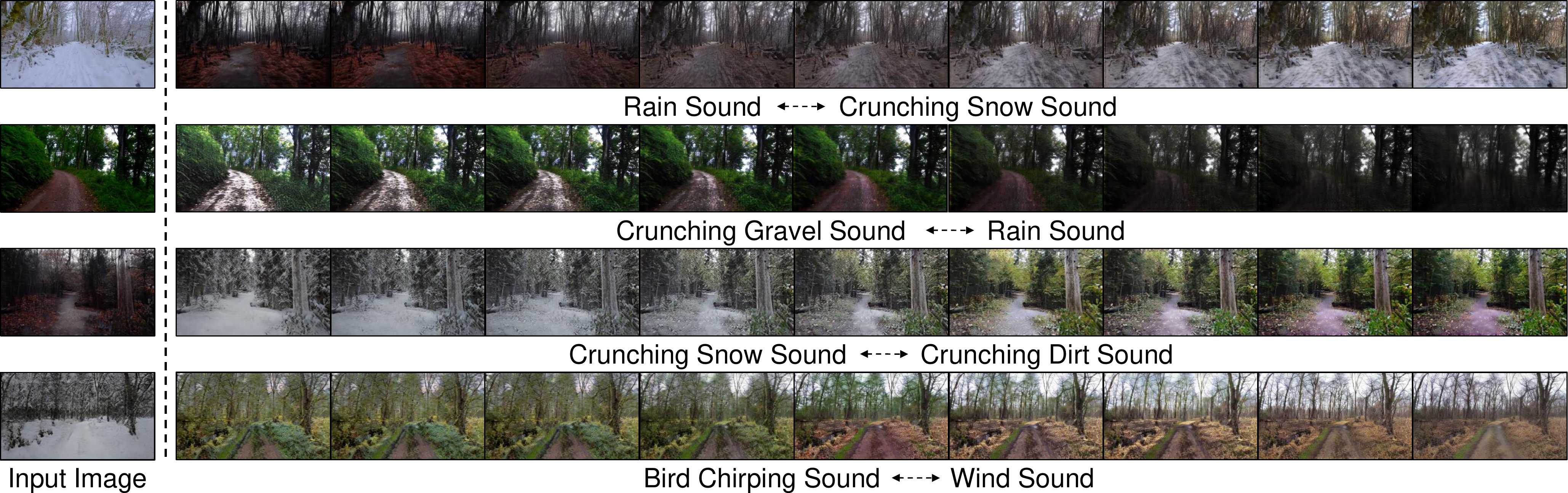}
	\caption{Qualitative results on image manipulation with different mixture sounds.} 
	\label{fig:audio_mixture}
\end{figure}

\mypar{Mixing sounds.}
We create sound mixtures by taking convex combinations of input sounds. The qualitative results are presented in Figure~\ref{fig:audio_mixture}. In the third row, for example, we can see that the snowy texture will be gradually erased while mixing a crunching snow sound with a muddy footstep sound from small to large. Furthermore, it appears to be a balanced state with both snowy and sunny features in the middle, \textit{i.e.}, white and green hues coexist. Surprisingly, such mixed audio is not available when our model is being trained. This linear additivity finding shows that audio cues have a prospective advantage over label ones for image translation.

\begin{figure}[t]
	\centering
	\includegraphics[width=\linewidth]{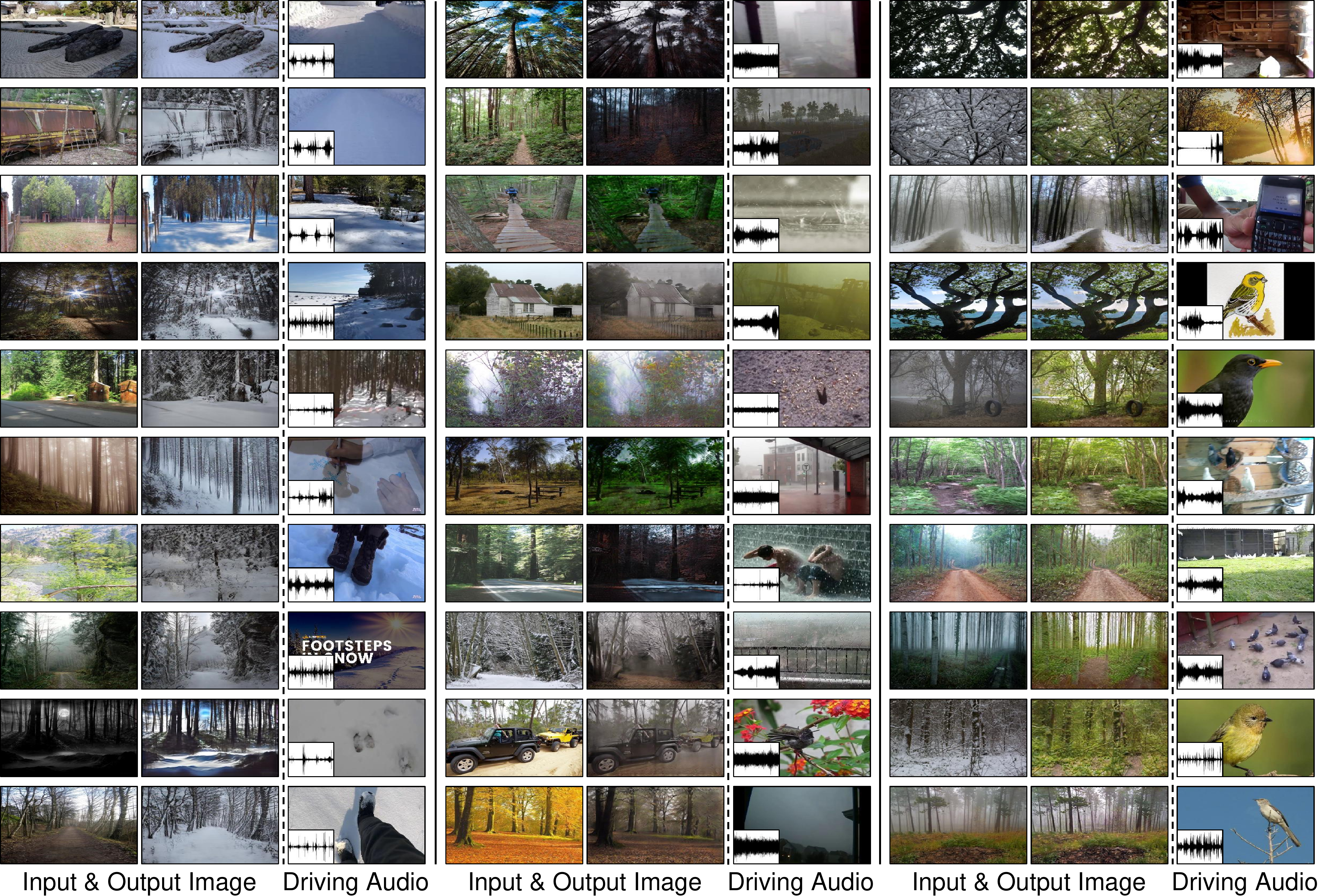}
	\caption{{\bf Qualitative generalization results.} We restyle images from Places~\cite{zhou2017places} using crunching snow and rain sounds taken from VGG-Sound~\cite{chen2020vggsound}.}
	\label{fig:generalization}
\end{figure}

\mypar{Generalization to other datasets.}
We ask whether our model can generalize to out-of-distribution data. We consider restyling images from the Places dataset \cite{zhou2017places} and audio from the VGG-Sound dataset \cite{chen2020vggsound} to examine our model's generalization performance. In Figure~\ref{fig:generalization}, we use crunching snow, rain and birds chirping sounds with a high probability of a class deduced by YAMNet~\cite{plakal2020yamnet}. Our model generates plausible images that match the content of in-the-wild audio.

\mypar{Adjusting an image's style through its sound.}
\begin{wrapfigure}[10]{r}{0.7\textwidth}
  \begin{center}
  \vspace{-11.5mm}
    \includegraphics[width=0.7\textwidth]{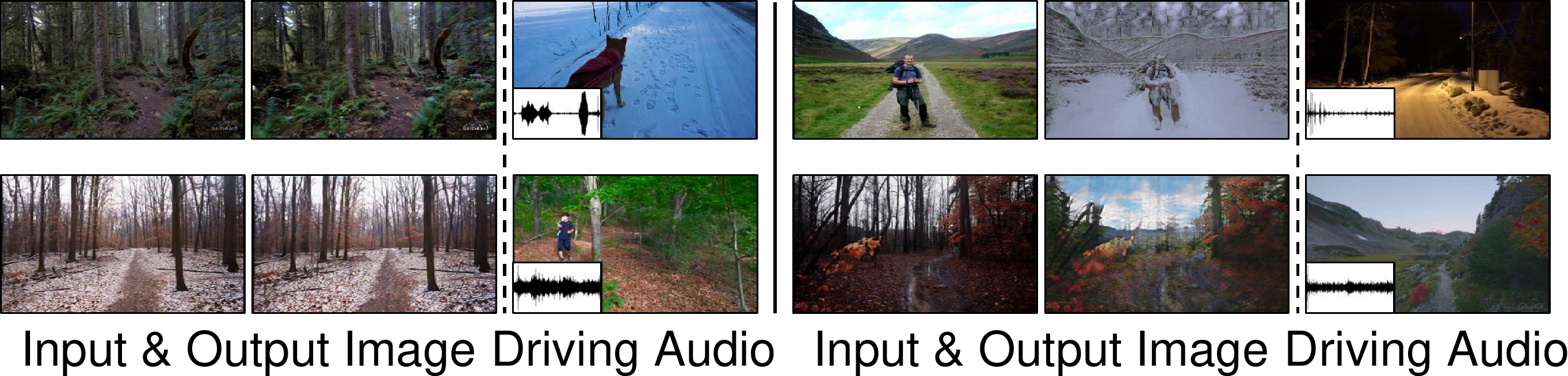}
  \end{center}
  \vspace{-5mm}
  \caption{{\bf Failure cases.} Our model fails to manipulate the style of the scene, perhaps due to the presence of speech in the sound (left). It also fails to learn how to style certain objects in a scene (right). }
  \label{fig:failure}
\end{wrapfigure}

We apply our method to a task inspired by video editing: adjusting an image's appearance by manipulating its {\em existing} sound. We take a video frame, manipulate its corresponding sound, and then resynthesize its video frames to match. This allows a user to make {\em consistent} changes to the two modalities: {\em e.g.}, an editor can adjust the volume of rain through intuitive volume-based controls, while automatically propagating these changes to images. 

\begin{figure}[t]
  \begin{center}
    \includegraphics[width=\textwidth]{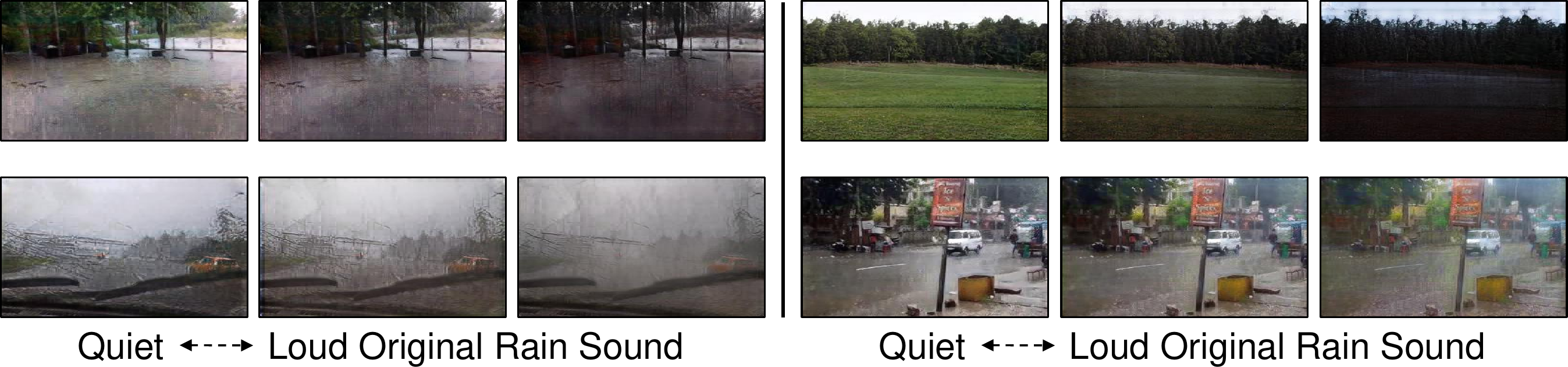}
  \end{center}
  \vspace{-3mm}
  \caption{{\bf Restyling with a video's existing sound.} We adjust the appearance of a video by increasing the volume of its soundtrack, and restyling the corresponding video frame.}
  \label{fig:application}
  \vspace{-3mm}
\end{figure}

We restyle videos from VGG-Sound~\cite{chen2020vggsound} by adjusting the volume of their already-existing soundtracks. Figure~\ref{fig:application} shows qualitative examples obtained by increasing the volume of videos recorded during light rain. As expected, the resulting images contains significantly more rain. 

\section{Discussion and Limitations}
Despite the fact that our model can yield promising results in various cases, the results are far from uniformly positive. Because ambient sounds in real life are diverse, our model can be easily upset with unexpected sounds. Figure~\ref{fig:failure} shows some typical failure cases. Specifically, if the sound is interfered by human speech, the learned translation will devolve to making minor adjustments to the input. As a result, handling a greater spectrum of mixture sound, particularly urban sound, will become increasingly important in the future. Another potential concern is that our model's performance will be suffered if the proportion of the scene to be converted is too small. In the lower right of Figure~\ref{fig:failure}, for example, the trees and sky each account for half of the input image, resulting in an odd conversion. This is because the model is unable to detect the region of the scene that needs conversion, but instead converts the entire scene. Nevertheless, as paired audio-visual data is ubiquitous in our daily life, this paper paves the way for image translation under the audio-visual context.
\section{Conclusion}
In this paper, we introduce a novel task called {\em audio-driven image stylization}, which aims to learn the visual styles from paired audio-visual data. To study this task, we propose a contrastive-based audio-visual GAN model, together with an unlabeled egocentric hiking dataset named \textit{Into the Wild}. Experimental results show that our model outperforms label and image conditioned baselines in both quantitative and qualitative evaluations. We also empirically find that changing the audio volume and mixture results in predictable visual changes. We hope our work will shed new light on cross-modal image synthesis.

\bibliographystyle{splncs04}
\bibliography{egbib}

\begin{thebibliography}{10}
\providecommand{\url}[1]{\texttt{#1}}
\providecommand{\urlprefix}{URL }
\providecommand{\doi}[1]{https://doi.org/#1}

\bibitem{afouras2018deep}
Afouras, T., Chung, J.S., Senior, A., Vinyals, O., Zisserman, A.: Deep
  audio-visual speech recognition. IEEE transactions on pattern analysis and
  machine intelligence  (2018)

\bibitem{arandjelovic2017look}
Arandjelovic, R., Zisserman, A.: Look, listen and learn. In: Proceedings of the
  IEEE International Conference on Computer Vision. pp. 609--617 (2017)

\bibitem{asano2020labelling}
Asano, Y.M., Patrick, M., Rupprecht, C., Vedaldi, A.: Labelling unlabelled
  videos from scratch with multi-modal self-supervision. In: Advances in Neural
  Information Processing Systems (2020)

\bibitem{bau2021paint}
Bau, D., Andonian, A., Cui, A., Park, Y., Jahanian, A., Oliva, A., Torralba,
  A.: Paint by word. In: arXiv:2103.10951 (2021)

\bibitem{chatterjee2020sound2sight}
Chatterjee, M., Cherian, A.: Sound2sight: Generating visual dynamics from sound
  and context. In: European Conference on Computer Vision. pp. 701--719.
  Springer (2020)

\bibitem{chen2021localizing}
Chen, H., Xie, W., Afouras, T., Nagrani, A., Vedaldi, A., Zisserman, A.:
  Localizing visual sounds the hard way. In: Proceedings of the Conference on
  Computer Vision and Pattern Recognition (CVPR) (2021)

\bibitem{chen2020vggsound}
Chen, H., Xie, W., Vedaldi, A., Zisserman, A.: Vggsound: A large-scale
  audio-visual dataset. In: ICASSP 2020-2020 IEEE International Conference on
  Acoustics, Speech and Signal Processing (ICASSP). pp. 721--725. IEEE (2020)

\bibitem{SimCLR}
Chen, T., Kornblith, S., Norouzi, M., Hinton, G.E.: A simple framework for
  contrastive learning of visual representations. In: International Conference
  on Machine Learning. pp. 1597--1607 (2020)

\bibitem{chen2021structure}
Chen, Z., Hu, X., Owens, A.: Structure from silence: Learning scene structure
  from ambient sound. In: 5th Annual Conference on Robot Learning (2021)

\bibitem{cheuk2020nnaudio}
Cheuk, K.W., Anderson, H., Agres, K., Herremans, D.: {nnAudio}: An on-the-fly
  gpu audio to spectrogram conversion toolbox using 1d convolutional neural
  networks. IEEE Access  \textbf{8},  161981--162003 (2020)

\bibitem{chung2017you}
Chung, J.S., Jamaludin, A., Zisserman, A.: You said that? In: British Machine
  Vision Conference (2017)

\bibitem{cramer2019look}
Cramer, J., Wu, H.H., Salamon, J., Bello, J.P.: Look, listen, and learn more:
  Design choices for deep audio embeddings. In: ICASSP 2019-2019 IEEE
  International Conference on Acoustics, Speech and Signal Processing (ICASSP).
  pp. 3852--3856. IEEE (2019)

\bibitem{dong2017semantic}
Dong, H., Yu, S., Wu, C., Guo, Y.: Semantic image synthesis via adversarial
  learning. In: Proceedings of the IEEE International Conference on Computer
  Vision. pp. 5706--5714 (2017)

\bibitem{ephrat2018looking}
Ephrat, A., Mosseri, I., Lang, O., Dekel, T., Wilson, K., Hassidim, A.,
  Freeman, W.T., Rubinstein, M.: Looking to listen at the cocktail party: A
  speaker-independent audio-visual model for speech separation. ACM
  Transactions on Graphics (TOG)  \textbf{37}(4) (2016)

\bibitem{fu2021language}
Fu, T.J., Wang, X.E., Wang, W.Y.: Language-driven image style transfer. arXiv
  preprint arXiv:2106.00178  (2021)

\bibitem{gan2020music}
Gan, C., Huang, D., Zhao, H., Tenenbaum, J.B., Torralba, A.: Music gesture for
  visual sound separation. In: Proceedings of the IEEE/CVF Conference on
  Computer Vision and Pattern Recognition. pp. 10478--10487 (2020)

\bibitem{gao2018learning}
Gao, R., Feris, R., Grauman, K.: Learning to separate object sounds by watching
  unlabeled video. In: Proceedings of the European Conference on Computer
  Vision (ECCV). pp. 35--53 (2018)

\bibitem{gao20192}
Gao, R., Grauman, K.: 2.5 d visual sound. In: Proceedings of the IEEE/CVF
  Conference on Computer Vision and Pattern Recognition. pp. 324--333 (2019)

\bibitem{gatys2015neural}
Gatys, L.A., Ecker, A.S., Bethge, M.: A neural algorithm of artistic style.
  arXiv preprint arXiv:1508.06576  (2015)

\bibitem{gemmeke2017audio}
Gemmeke, J.F., Ellis, D.P., Freedman, D., Jansen, A., Lawrence, W., Moore,
  R.C., Plakal, M., Ritter, M.: Audio set: An ontology and human-labeled
  dataset for audio events. In: 2017 IEEE International Conference on
  Acoustics, Speech and Signal Processing (ICASSP). pp. 776--780. IEEE (2017)

\bibitem{ginosar2019learning}
Ginosar, S., Bar, A., Kohavi, G., Chan, C., Owens, A., Malik, J.: Learning
  individual styles of conversational gesture. In: Proceedings of the IEEE/CVF
  Conference on Computer Vision and Pattern Recognition. pp. 3497--3506 (2019)

\bibitem{adversarial_loss}
Goodfellow, I., Pouget-Abadie, J., Mirza, M., Xu, B., Warde-Farley, D., Ozair,
  S., Courville, A., Bengio, Y.: Generative adversarial networks. In: Advances
  in Neural Information Processing Systems. pp. 2672--2680 (2014)

\bibitem{NCE}
Gutmann, M., Hyv{\"a}rinen, A.: Noise-contrastive estimation: A new estimation
  principle for unnormalized statistical models. In: Proceedings of the
  Thirteenth International Conference on Artificial Intelligence and
  Statistics. pp. 297--304 (2010)

\bibitem{harwath2018jointly}
Harwath, D., Recasens, A., Sur{\'\i}s, D., Chuang, G., Torralba, A., Glass, J.:
  Jointly discovering visual objects and spoken words from raw sensory input.
  In: Proceedings of the European conference on computer vision (ECCV). pp.
  649--665 (2018)

\bibitem{he2016deep}
He, K., Zhang, X., Ren, S., Sun, J.: Deep residual learning for image
  recognition. In: Proceedings of the IEEE conference on computer vision and
  pattern recognition. pp. 770--778 (2016)

\bibitem{hershey2017cnn}
Hershey, S., Chaudhuri, S., Ellis, D.P., Gemmeke, J.F., Jansen, A., Moore,
  R.C., Plakal, M., Platt, D., Saurous, R.A., Seybold, B., et~al.: Cnn
  architectures for large-scale audio classification. In: 2017 ieee
  international conference on acoustics, speech and signal processing (icassp).
  pp. 131--135. IEEE (2017)

\bibitem{hertzmann2001image}
Hertzmann, A., Jacobs, C.E., Oliver, N., Curless, B., Salesin, D.H.: Image
  analogies. In: Proceedings of the 28th annual conference on Computer graphics
  and interactive techniques. pp. 327--340 (2001)

\bibitem{heusel2017gans}
Heusel, M., Ramsauer, H., Unterthiner, T., Nessler, B., Hochreiter, S.: Gans
  trained by a two time-scale update rule converge to a local nash equilibrium.
  In: Advances in Neural Information Processing Systems (2017)

\bibitem{hu2021neural}
Hu, C., Tian, Q., Li, T., Wang, Y., Wang, Y., Zhao, H.: Neural dubber: Dubbing
  for silent videos according to scripts. In: Advances in neural information
  processing systems (2021)

\bibitem{huang2017arbitrary}
Huang, X., Belongie, S.: Arbitrary style transfer in real-time with adaptive
  instance normalization. In: Proceedings of the IEEE International Conference
  on Computer Vision. pp. 1501--1510 (2017)

\bibitem{iashin2021taming}
Iashin, V., Rahtu, E.: Taming visually guided sound generation. arXiv preprint
  arXiv:2110.08791  (2021)

\bibitem{isola2017image}
Isola, P., Zhu, J.Y., Zhou, T., Efros, A.A.: Image-to-image translation with
  conditional adversarial networks. In: Proceedings of the IEEE conference on
  computer vision and pattern recognition. pp. 1125--1134 (2017)

\bibitem{johnson2016perceptual}
Johnson, J., Alahi, A., Fei-Fei, L.: Perceptual losses for real-time style
  transfer and super-resolution. In: European conference on computer vision.
  pp. 694--711. Springer (2016)

\bibitem{johnson2018image}
Johnson, J., Gupta, A., Fei-Fei, L.: Image generation from scene graphs. In:
  Proceedings of the IEEE conference on computer vision and pattern
  recognition. pp. 1219--1228 (2018)

\bibitem{kim2017learning}
Kim, T., Cha, M., Kim, H., Lee, J.K., Kim, J.: Learning to discover
  cross-domain relations with generative adversarial networks. In:
  International Conference on Machine Learning. pp. 1857--1865. PMLR (2017)

\bibitem{kingma2014adam}
Kingma, D.P., Ba, J.: Adam: A method for stochastic optimization. In:
  International Conference for Learning Representations (2015)

\bibitem{korbar2018cooperative}
Korbar, B., Tran, D., Torresani, L.: Cooperative learning of audio and video
  models from self-supervised synchronization. In: Proceedings of the Advances
  in Neural Information Processing Systems (2018)

\bibitem{laffont2014transient}
Laffont, P.Y., Ren, Z., Tao, X., Qian, C., Hays, J.: Transient attributes for
  high-level understanding and editing of outdoor scenes. ACM Transactions on
  graphics (TOG)  \textbf{33}(4),  1--11 (2014)

\bibitem{langlois2014inverse}
Langlois, T.R., James, D.L.: Inverse-foley animation: Synchronizing rigid-body
  motions to sound. ACM Transactions on Graphics (TOG)  \textbf{33}(4),  1--11
  (2014)

\bibitem{lee2021sound}
Lee, S.H., Roh, W., Byeon, W., Yoon, S.H., Kim, C.Y., Kim, J., Kim, S.:
  Sound-guided semantic image manipulation. arXiv preprint arXiv:2112.00007
  (2021)

\bibitem{levine2010gesture}
Levine, S., Kr{\"a}henb{\"u}hl, P., Thrun, S., Koltun, V.: Gesture controllers.
  In: ACM SIGGRAPH. pp. 1--11 (2010)

\bibitem{mahajan2018exploring}
Mahajan, D., Girshick, R., Ramanathan, V., He, K., Paluri, M., Li, Y.,
  Bharambe, A., Van Der~Maaten, L.: Exploring the limits of weakly supervised
  pretraining. In: Proceedings of the European conference on computer vision
  (ECCV). pp. 181--196 (2018)

\bibitem{mao2017least}
Mao, X., Li, Q., Xie, H., Lau, R.Y., Wang, Z., Paul~Smolley, S.: Least squares
  generative adversarial networks. In: Proceedings of the IEEE international
  conference on computer vision. pp. 2794--2802 (2017)

\bibitem{mikolov2013efficient}
Mikolov, T., Chen, K., Corrado, G., Dean, J.: Efficient estimation of word
  representations in vector space. arXiv preprint arXiv:1301.3781  (2013)

\bibitem{morgado2018self}
Morgado, P., Vasconcelos, N., Langlois, T., Wang, O.: Self-supervised
  generation of spatial audio for 360 video. In: Advances in Neural Information
  Processing Systems (2018)

\bibitem{morgado2021audio}
Morgado, P., Vasconcelos, N., Misra, I.: Audio-visual instance discrimination
  with cross-modal agreement. In: Proceedings of the IEEE/CVF Conference on
  Computer Vision and Pattern Recognition. pp. 12475--12486 (2021)

\bibitem{nam2018text}
Nam, S., Kim, Y., Kim, S.J.: Text-adaptive generative adversarial networks:
  Manipulating images with natural language. In: Advances in neural information
  processing systems (2018)

\bibitem{ngiam2011multimodal}
Ngiam, J., Khosla, A., Kim, M., Nam, J., Lee, H., Ng, A.Y.: Multimodal deep
  learning. In: ICML (2011)

\bibitem{owens2018audio}
Owens, A., Efros, A.A.: Audio-visual scene analysis with self-supervised
  multisensory features. In: Proceedings of the European Conference on Computer
  Vision (2018)

\bibitem{owens2016visually}
Owens, A., Isola, P., McDermott, J., Torralba, A., Adelson, E.H., Freeman,
  W.T.: Visually indicated sounds. In: Proceedings of the IEEE conference on
  computer vision and pattern recognition. pp. 2405--2413 (2016)

\bibitem{park2020contrastive}
Park, T., Efros, A.A., Zhang, R., Zhu, J.Y.: Contrastive learning for unpaired
  image-to-image translation. In: European Conference on Computer Vision. pp.
  319--345 (2020)

\bibitem{parmar2021buggy}
Parmar, G., Zhang, R., Zhu, J.Y.: On buggy resizing libraries and surprising
  subtleties in fid calculation. arXiv preprint arXiv:2104.11222  (2021)

\bibitem{plakal2020yamnet}
Plakal, M., Ellis, D.: {YAMNet}. Jan 2020 [Online], available:
  \url{https://github.com/tensorflow/models/tree/master/research/audioset/yamnet}

\bibitem{prajwal2020learning}
Prajwal, K., Mukhopadhyay, R., Namboodiri, V.P., Jawahar, C.: Learning
  individual speaking styles for accurate lip to speech synthesis. In:
  Proceedings of the IEEE/CVF Conference on Computer Vision and Pattern
  Recognition. pp. 13796--13805 (2020)

\bibitem{prajwal2020lip}
Prajwal, K., Mukhopadhyay, R., Namboodiri, V.P., Jawahar, C.: A lip sync expert
  is all you need for speech to lip generation in the wild. In: Proceedings of
  the 28th ACM International Conference on Multimedia. pp. 484--492 (2020)

\bibitem{radford2021learning}
Radford, A., Kim, J.W., Hallacy, C., Ramesh, A., Goh, G., Agarwal, S., Sastry,
  G., Askell, A., Mishkin, P., Clark, J., et~al.: Learning transferable visual
  models from natural language supervision. In: International conference on
  machine learning (2021)

\bibitem{ramesh2021zero}
Ramesh, A., Pavlov, M., Goh, G., Gray, S., Voss, C., Radford, A., Chen, M.,
  Sutskever, I.: Zero-shot text-to-image generation. arXiv preprint
  arXiv:2102.12092  (2021)

\bibitem{reed2016generative}
Reed, S., Akata, Z., Yan, X., Logeswaran, L., Schiele, B., Lee, H.: Generative
  adversarial text to image synthesis. In: International Conference on Machine
  Learning. pp. 1060--1069 (2016)

\bibitem{de1994learning}
de~Sa, V.R.: Learning classification with unlabeled data. In: Advances in
  neural information processing systems. pp. 112--119. Citeseer (1994)

\bibitem{shlizerman2018audio}
Shlizerman, E., Dery, L., Schoen, H., Kemelmacher-Shlizerman, I.: Audio to body
  dynamics. In: Proceedings of the IEEE conference on computer vision and
  pattern recognition. pp. 7574--7583 (2018)

\bibitem{szegedy2016rethinking}
Szegedy, C., Vanhoucke, V., Ioffe, S., Shlens, J., Wojna, Z.: Rethinking the
  inception architecture for computer vision. In: Proceedings of the IEEE
  conference on computer vision and pattern recognition. pp. 2818--2826 (2016)

\bibitem{tenenbaum2000separating}
Tenenbaum, J.B., Freeman, W.T.: Separating style and content with bilinear
  models. Neural computation  \textbf{12}(6),  1247--1283 (2000)

\bibitem{wang2020makes}
Wang, W., Tran, D., Feiszli, M.: What makes training multi-modal classification
  networks hard? In: Proceedings of the IEEE/CVF Conference on Computer Vision
  and Pattern Recognition. pp. 12695--12705 (2020)

\bibitem{wu2020describing}
Wu, C., Timm, M., Maji, S.: Describing textures using natural language. In:
  Computer Vision--ECCV 2020: 16th European Conference, Glasgow, UK, August
  23--28, 2020, Proceedings, Part I 16. pp. 52--70. Springer (2020)

\bibitem{wu2021wav2clip}
Wu, H.H., Seetharaman, P., Kumar, K., Bello, J.P.: Wav2clip: Learning robust
  audio representations from clip. arXiv preprint arXiv:2110.11499  (2021)

\bibitem{yang2020telling}
Yang, K., Russell, B., Salamon, J.: Telling left from right: Learning spatial
  correspondence of sight and sound. In: Proceedings of the IEEE/CVF Conference
  on Computer Vision and Pattern Recognition. pp. 9932--9941 (2020)

\bibitem{yi2017dualgan}
Yi, Z., Zhang, H., Tan, P., Gong, M.: Dualgan: Unsupervised dual learning for
  image-to-image translation. In: Proceedings of the IEEE international
  conference on computer vision. pp. 2849--2857 (2017)

\bibitem{zhang2017generative}
Zhang, Z., Wu, J., Li, Q., Huang, Z., Traer, J., McDermott, J.H., Tenenbaum,
  J.B., Freeman, W.T.: Generative modeling of audible shapes for object
  perception. In: Proceedings of the IEEE International Conference on Computer
  Vision. pp. 1251--1260 (2017)

\bibitem{zhao2019sound}
Zhao, H., Gan, C., Ma, W.C., Torralba, A.: The sound of motions. In:
  Proceedings of the IEEE/CVF International Conference on Computer Vision. pp.
  1735--1744 (2019)

\bibitem{zhao2018sound}
Zhao, H., Gan, C., Rouditchenko, A., Vondrick, C., McDermott, J., Torralba, A.:
  The sound of pixels. In: Proceedings of the European conference on computer
  vision (ECCV). pp. 570--586 (2018)

\bibitem{zhou2017places}
Zhou, B., Lapedriza, A., Khosla, A., Oliva, A., Torralba, A.: Places: A 10
  million image database for scene recognition. IEEE transactions on pattern
  analysis and machine intelligence  \textbf{40}(6),  1452--1464 (2017)

\bibitem{zhou2019talking}
Zhou, H., Liu, Y., Liu, Z., Luo, P., Wang, X.: Talking face generation by
  adversarially disentangled audio-visual representation. In: Proceedings of
  the AAAI Conference on Artificial Intelligence. pp. 9299--9306 (2019)

\bibitem{zhou2018visual}
Zhou, Y., Wang, Z., Fang, C., Bui, T., Berg, T.L.: Visual to sound: Generating
  natural sound for videos in the wild. In: Proceedings of the IEEE Conference
  on Computer Vision and Pattern Recognition. pp. 3550--3558 (2018)

\bibitem{zhu2017unpaired}
Zhu, J.Y., Park, T., Isola, P., Efros, A.A.: Unpaired image-to-image
  translation using cycle-consistent adversarial networks. In: Proceedings of
  the IEEE conference on computer vision and pattern recognition. pp.
  2223--2232 (2017)

\end{thebibliography}
\clearpage
\appendix
\section{Appendix}
\label{sec: appendix}

\vspace{-2mm}
\subsection{\textbf{\textit{Into the Wild}} dataset}
\label{sec: appendix_dataset}
We introduce the \textit{Into the Wild} dataset, a set of egocentric hiking videos for our proposed audio-driven image stylization (ADIS), because hiking is featured with a strong audio-visual association of nature.

We collected these videos on YouTube by searching for the keywords like hike+POV, hike+footsteps, hike+ASMR, and hike+binaural. We employ YAMNet \cite{plakal2020yamnet} to tag each associated soundtrack to ensure that they play the actual sound and are not replaced by any other sounds, such as background music.

\begin{figure}[t]
	\centering
	\begin{subfigure}[h]{0.48\linewidth}
		\centering
		\includegraphics[width=0.8\linewidth]{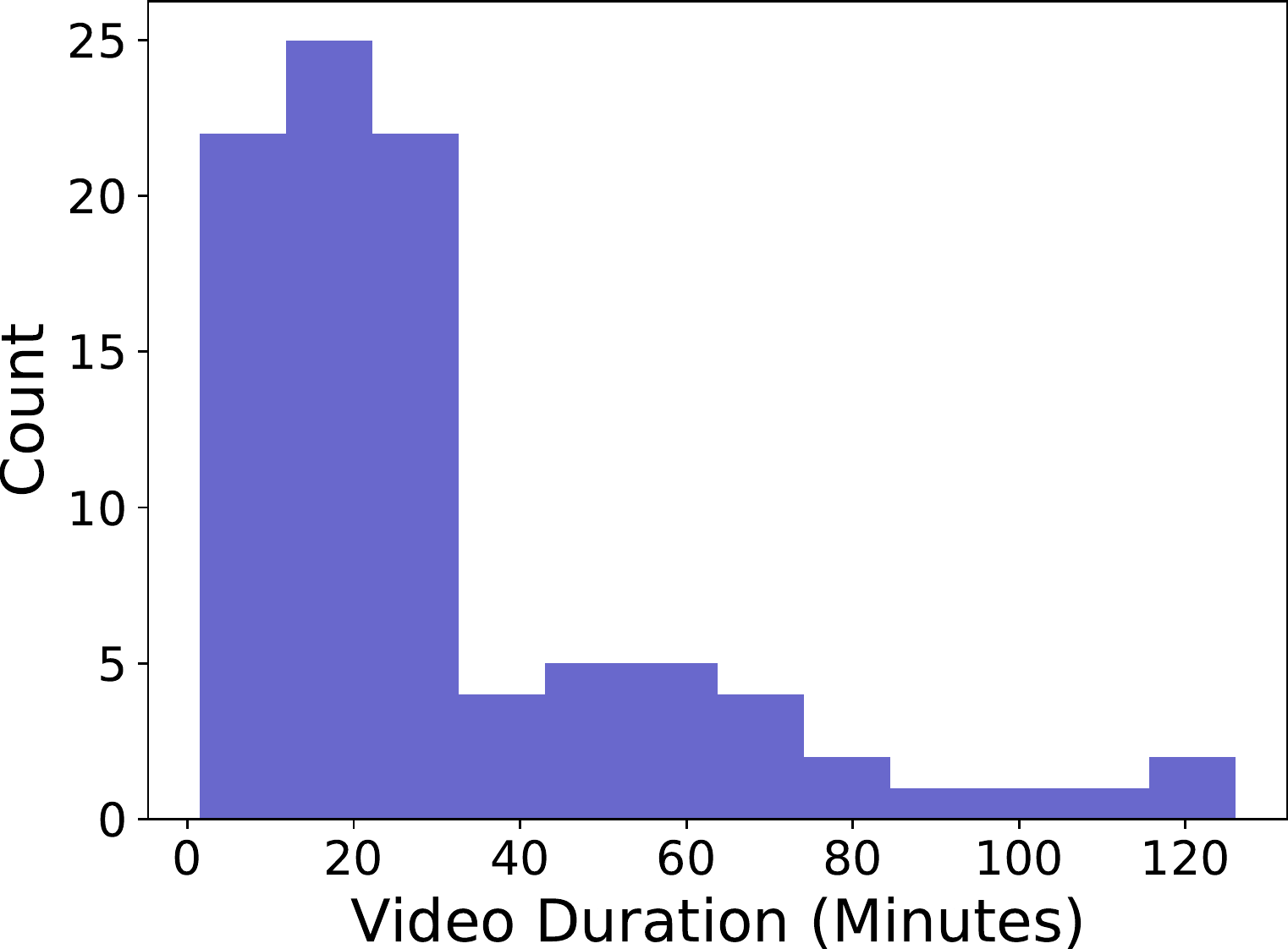}
		\caption{Distribution of video duration}
		\label{fig:duration}
	\end{subfigure}
	\begin{subfigure}[h]{0.48\linewidth}
		\centering
		\includegraphics[width=0.78\linewidth]{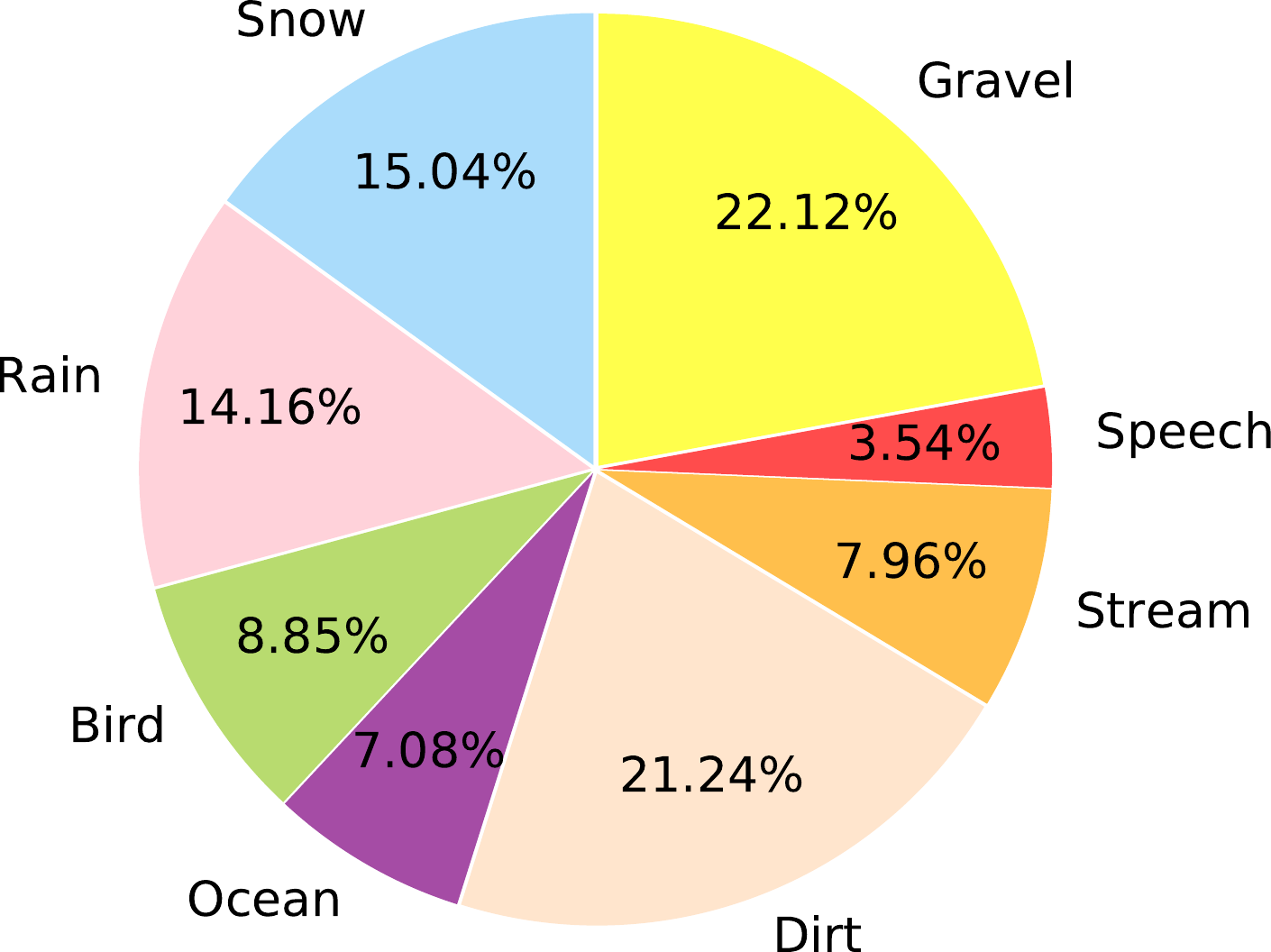}
		\caption{Distribution of video categories}
		\label{fig:categories}
	\end{subfigure}
	\caption{Statistical analysis of the \textit{Into the Wild} Dataset.}
	\label{fig:dataset}
\end{figure}

The duration statistics of the \textit{Into the Wild} dataset are shown in Figure~\ref{fig:duration}. Specifically, it contains 94 untrimmed videos, some of which are already presented in Figure 4 of the main paper. Please note that the category labels of these videos are not labeled by humans, but acquired from the YAMNet \cite{plakal2020yamnet} predictions, which roughly consist of 8 categories: crunching snow, gravel, and dirt; rain; birds chirping; ocean; stream and human speech. The detailed categorical distribution is illustrated in Figure~\ref{fig:categories}. % ~\ref{fig:example_frame}

\appsubsec{Training Details}
\label{sec: appendix_details}
\paragraph{{\bf Training Setting}}
Except for the batch size and audio network, we intentionally match the architecture and hyperparameter settings with CycleGAN \cite{zhu2017unpaired} and CUT \cite{park2020contrastive}. We employ ResNet-based generator \cite{johnson2016perceptual} with 9 residual blocks, PatchGAN discriminator \cite{isola2017image}, Least Square GAN loss \cite{mao2017least}, ResNet18-based audio encoder \cite{he2016deep}, with the batch size of 16, and the Adam optimizer \cite{kingma2014adam} with 0.002 learning rate. Both $\lambda$ and $\mu$ in Eq.(4) of the main paper are set to 0.5.

Our model is trained for 50 epochs, with the learning rate remaining constant for the first 30 epochs and linearly decaying to zero over the last 20 epochs. The encoder $G_\text{enc}$ follows the first half of the CycleGAN generator \cite{zhu2017unpaired}. We also extract features from 5 different scales to calculate the patch-based structure discriminator loss: the input RGB pixels, the first and second downsampling convolution features, and the first and fifth residual block features. We sample 256 random locations for each layer's features and apply a 2-layer MLP to obtain 256-dimension features as the final output for computing the multi-scale patch-wise contrastive loss.

\apppar{\textit{Into the Wild} dataset}
We divide all of the videos into 3-seconds video clips, then uniformly sample 8 frames from each video clip to save as images, yielding a total of 454560 images and 56820 audios. We then randomly sample 20\% audios as the test set.

\apppar{The \textit{Greatest Hits} dataset}
We first identify the videos by the type of object being hit on, and then only the outdoor videos are used for training: dirt, grass, gravel, leaf, and water, resulting in a total of 32172 images and 8043 audios. We then select 15\% audios at random as the test set.

\begin{figure}[t]
	\centering
	\includegraphics[width=0.88\linewidth]{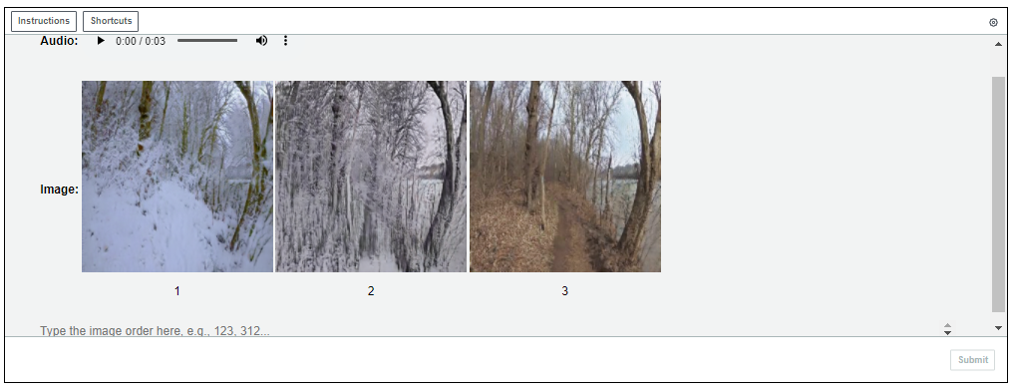}
	\caption{A screenshot of AMT for rating the audio-visual correspondence.}
	\label{fig:mturk}
\end{figure}

\appsubsec{Evaluation Details}
\label{sec: appendix_evaluation}
\paragraph{{\bf Audio-visual Correspondence (AVC)}} A two-stream network is utilized to compute AVC \cite{arandjelovic2017look}, with one stream extracting audio feature and the other extracting visual feature. Specifically, we apply OpenL3 \cite{cramer2019look} to obtain these features, and then compute the average cosine similarity for each image-audio pair. To be more explicit, we employ an ``env" content type pre-trained model with 512-dimensional linear spectrogram representation.

\apppar{Fréchet Inception Distance (FID)} FID \cite{heusel2017gans} is calculated by scaling the images to 299-by-299 using the PyTorch framework's bi-linear sampling, and then take the activation of the last average pooling layer of a pre-trained Inception V3 \cite{szegedy2016rethinking}. We adopt Clean-FID \cite{parmar2021buggy} to circumvent the issue that FID computation requires complicated and error-prone steps, such as the resizing functions in different libraries often produce inaccurate results.

\apppar{Contrastive Language-Image Pre-Training (CLIP)} \cite{radford2021learning} is computed by performing contrastive pre-training on a variety of image-text pairs. It's widely known for zero-shot prediction, but we use it as a feature extractor to compute the cosine similarity between images and labels in order to assess conversion quality. To calculate it, we leverage an off-the-shelf ``ViT-B/32" CLIP model \cite{radford2021learning}.

\apppar{Amazon Mechanical Turk (AMT)} In addition to the objective evaluations mentioned above, we employ AMT to study the relationship between audio and visual from a subjective standpoint, \textit{i.e.}, human perspective. A screenshot of the demo page is shown in Figure~\ref{fig:mturk}. The MTurker is required to rank such correlations based on audios and images generated by our method and the baseline methods, with the best earning 4 points and the worst earning 1 point. Thus, the scores range from 1 to 4. Notably, twenty Mturkers were asked to rank a total of 1000 random samples from the test set in our case. The final scores are reported on average.

\begin{table}[t]
	\centering
	\caption{Quantitative comparison for different pre-training methods on the \textit{Into the Wild} dataset.}
     \setlength{\tabcolsep}{2.95mm}{
	\begin{tabular}{l c c | c}
		\toprule
    	\multirow{2}*{\textbf{Pre-training Method}} & \multicolumn{3}{c}{\textbf{Objective Evaluation}} \\ 
		\cmidrule{2-4} & AVC ($\uparrow$) & FID ($\downarrow$) & CLIP ($\uparrow$) \\
		\midrule
		Ours (from scratch) & 0.820 & 34.139 & 0.238 \\
		+ SeLaVi~\cite{asano2020labelling} & 0.822 & 32.882 & 0.242 \\
		+ Wav2CLIP~\cite{wu2021wav2clip} & \textbf{0.831} & \textbf{30.334} & \textbf{0.246} \\
		\bottomrule
	\end{tabular}}
	\label{tb:pre-train}
\end{table}

\begin{figure}[t]
	\centering
	\includegraphics[width=\linewidth]{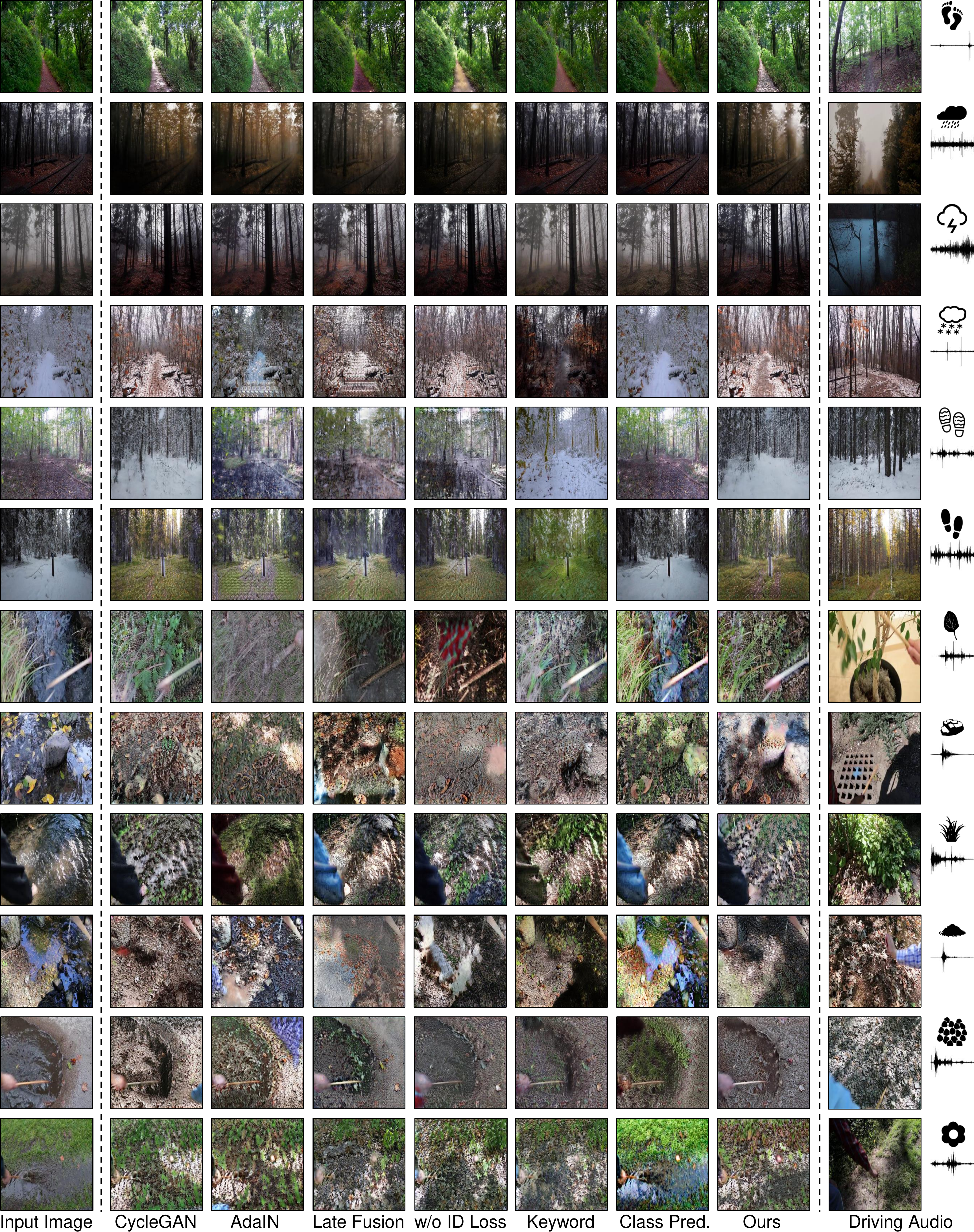}
	\caption{Randomly selected qualitative results of our model, baselines and ablations. This is an extension of Figure~\ref{fig:results} in the main paper.}
	\label{fig:additional_quality}
\end{figure}

\appsubsec{Additional Results}
\label{sec: appendix_additional_results}
\paragraph{{\bf Additional qualitative comparisons}}
Additional qualitative comparisons on our method to the baselines and ablations are shown in Figure~\ref{fig:additional_quality}. It turns out that our model produces better or competitive results, exhibiting its versatility compared to label-based baselines.

\begin{figure}[t]
	\centering
	\includegraphics[width=\linewidth]{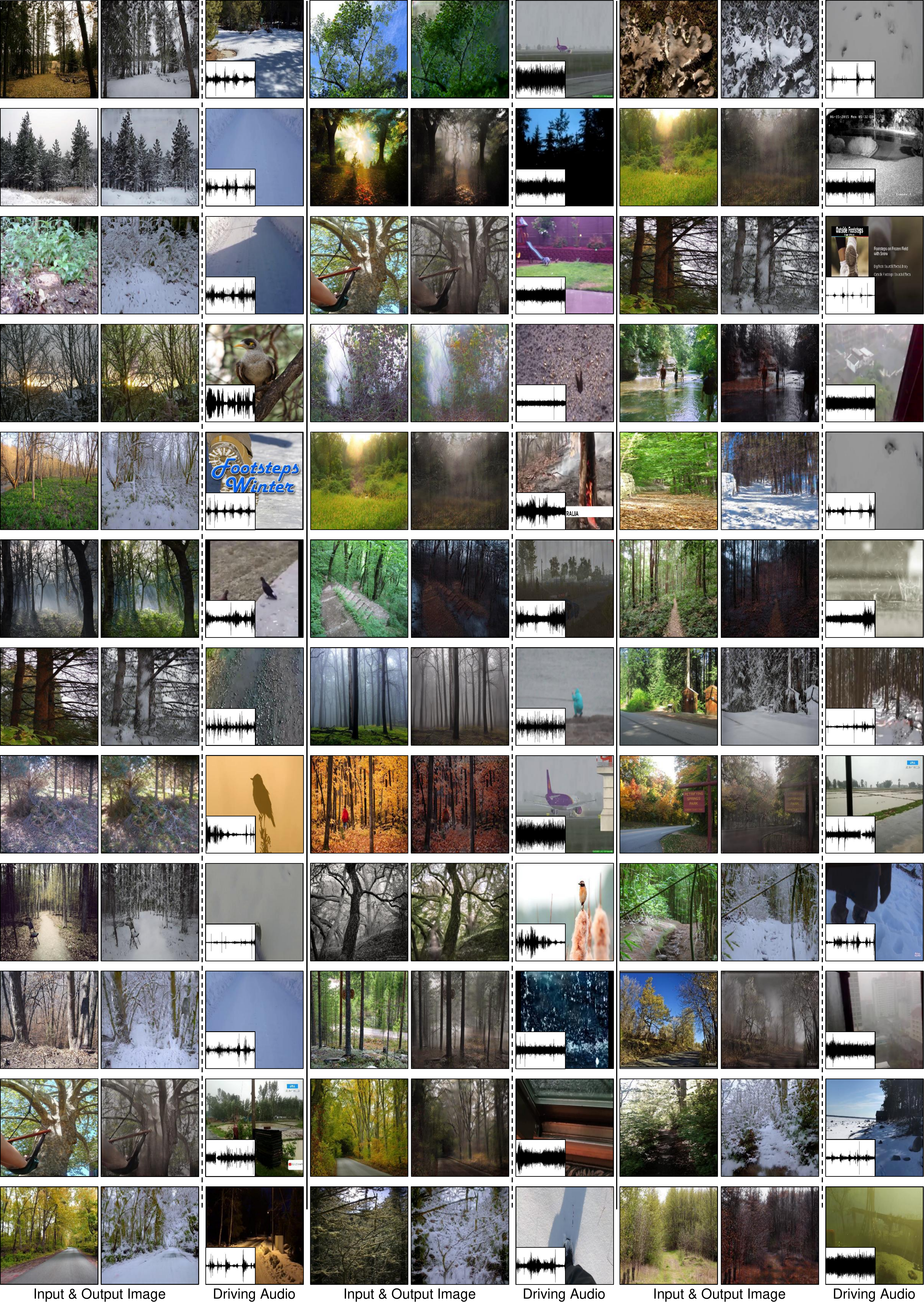}
	\caption{Randomly selected qualitative results of generalization experiment. This is an extension of Figure~\ref{fig:generalization} in the main paper.}
	\label{fig:additional_generalization}
\end{figure}

\apppar{Additional generalization results}
Additional qualitative results of the generalization experiment are shown in Figure~\ref{fig:additional_generalization}. These are accomplished by using images from the Places dataset \cite{zhou2017places} and the audios from the VGG-Sound dataset \cite{chen2020vggsound}. Our model is able to generate plausible images that match the content of the out-of-distribution audio.
 
\apppar{Additional pre-training comparisons}
We also use Wav2CLIP \cite{wu2021wav2clip}, an audio representation learning method derived on CLIP \cite{radford2021learning}, to fine-tune ADIS. To transfer knowledge, it employs a frozen image model to bridge the gap between a sophisticated language model and a scratch audio model. Wav2CLIP could be a better pre-training method for ADIS than SeLaVi \cite{asano2020labelling} since it is implicitly exposed to numerous well-annotated image-text pairs. Table~\ref{tb:pre-train} shows the quantitative comparison results. It appears that Wav2CLIP surpasses both training from scratch and SeLaVi pre-training methods with respect to the AVC, FID, and CLIP metrics, indicating that it has a stronger representation ability than the others.

\newpage

\newpage

\end{document}